\documentclass[acmsmall,nonacm]{acmart}

\usepackage{graphicx}%
\usepackage{multirow}%
\usepackage{amsthm}%
\usepackage{mathrsfs}%
\usepackage[title]{appendix}%
\usepackage{xcolor}%
\usepackage{textcomp}%
\usepackage{manyfoot}%
\usepackage{booktabs}%
\usepackage{algorithm}%
\usepackage{algorithmic}%
\usepackage{listings}%
\usepackage{subcaption}
\usepackage{caption}
\usepackage{enumitem}

\newtheorem{example}{Example}[section]
\usepackage{soul}

\newcommand{\acronym}{ADL-LLM} 

\AtBeginDocument{%
  }

\setcopyright{acmlicensed}
\copyrightyear{2018}
\acmYear{2018}
\acmDOI{XXXXXXX.XXXXXXX}

\acmJournal{TIST}
\acmVolume{37}
\acmNumber{4}
\acmArticle{111}
\acmMonth{8}

\begin{document}

\title{Large Language Models are Zero-Shot Recognizers for Activities of Daily Living}

\author{Gabriele Civitarese}
\orcid{0000-0002-8247-2524}
\email{gabriele.civitarese@unimi.it}
\author{Michele Fiori}
\orcid{0009-0000-2462-3075}
\email{michele.fiori@unimi.it}
\author{Priyankar Choudhary}
\orcid{0000-0003-2688-7415}
\email{priyankar.choudhary@unimi.it}
\author{Claudio Bettini}
\orcid{0000-0002-1727-7650}
\email{claudio.bettini@unimi.it}
\affiliation{%
  \institution{EveryWare Lab, Dept. of Computer Science, University of Milan}
  \country{Italy}
}

\renewcommand{\shortauthors}{Civitarese et al.}

\begin{abstract}
The sensor-based recognition of Activities of Daily Living (ADLs) in smart home environments enables several applications in the areas of energy management, safety, well-being, and healthcare.
ADL recognition is typically based on deep learning methods requiring large datasets to be trained.
Recently, several studies proved that Large Language Models (LLMs) effectively capture common-sense knowledge about human activities. However, the effectiveness of LLMs for ADL recognition in smart home environments still deserves to be investigated. In this work, we propose \acronym{}, a novel LLM-based ADL recognition system. \acronym{} transforms 
raw sensor data into textual representations, that are processed by an LLM to perform zero-shot ADL recognition.
Moreover, in the scenario where a small labeled dataset is available, \acronym{} can also be empowered with few-shot prompting. We evaluated \acronym{} on two public datasets, showing its effectiveness in this domain.
\end{abstract}




\maketitle

\textit{\textbf{\textcolor{red}{Paper accepted for publication in the ACM Transactions on Intelligent Systems and Technology (TIST) journal.}
}}

\section{Introduction}

Sensor-based Human Activity Recognition (HAR) is a well-known research topic extensively studied in pervasive computing~\cite{chen2021deep}.
The HAR task involves determining the most likely activity performed by a subject through the intelligent analysis of sensor data collected from one or multiple devices (e.g., wearables, environmental, radio-based). The adopted sensing technologies influence the granularity at which activities can be recognized. For instance, using inertial sensors in mobile/wearable devices makes it possible to detect physical activities like running, walking, and standing still. On the other hand, environmental sensors are particularly useful for recognizing high-level Activities of Daily Living (ADLs) like cooking, taking medicines, or watering plants in smart-home environments, also considering that subjects at home are less inclined to carry or wear devices all the time. 

Among the many applications of sensor-based ADL recognition in the areas of energy management, safety, well-being, and healthcare, the long-term monitoring of the ADLs performed by elderly subjects in their homes gained significant importance~\cite{rashidi2012survey}.
The continuous advances in the medical domain significantly extended the world population's life expectancy~\cite{chen2023digital}. While this is surely an outstanding achievement, aging also brings several challenges to society. Among the aging-related issues, cognitive decline is one of the greatest health threats of old age, with nearly $50\%$ of adults over the age of $85$ afflicted with Alzheimer’s disease~\cite{bishop2010neural}. Detecting the early symptoms of cognitive decline is crucial to implementing primary or secondary preventive therapeutic strategies~\cite{hudon2020behavioral}. 
One of the main markers used by clinicians to assess the cognitive status of a subject is represented by ADLs, whose impairment (i.e., difficulty or inability to perform them independently) is correlated to a rapid functional decline and a higher likelihood of progression to dementia~\cite{jefferson2008characterization}.
However, the subjects at risk usually have only a few sporadic visits with medical experts (e.g., neurologists, neuropsychologists, and geriatricians). Such visits often fail to accurately diagnose cognitive decline, especially when it may be revealed by subtle behavioral changes. Indeed, besides a few instrumental tests, and neuro-psychological evaluations at the time of visit, the clinicians mainly rely on possibly inaccurate reports provided by the subjects themselves or their caregivers. It is hence of great value to implement smart telemedicine strategies to continuously and unobtrusively monitor the subjects at risk while keeping them in the comfort of their homes.

In the last decades, several research groups proposed the use of unobtrusive sensing infrastructures in smart homes to continuously recognize ADLs in smart homes, to identify digital markers that may indicate the onset of cognitive decline~\cite{khodabandehloo2021healthxai,grammatikopoulou2024assessing}.
For the sake of privacy, most of the existing solutions do not consider cameras or microphones (that may not be accepted in home environments), but they are usually based on environmental sensors monitoring the interaction of the subject with the home infrastructure (e.g., motion sensors, door sensors, pressure sensors).

However, the majority of the proposed approaches to automatically detect ADLs from sensor data require training deep learning models~\cite{chen2021deep}, thus necessitating the collection of large datasets in the home environment (possibly labeled with the ground truth reporting what activity is being performed). Labeled data collection in these settings is privacy-intrusive, costly, and prohibitive~\cite{babangida2022internet}. Another well-known line of research in this domain considers the adoption of purely knowledge-based approaches~\cite{chen2009ontology}. The rationale behind these methods is that common-sense knowledge about human activities (e.g., a person usually prepares their meals in the kitchen, turns on the stove, and uses cooking instruments) can be modeled with a set of rules. Knowledge-based ADL recognition is performed by mapping sensor data windows to semantic concepts and performing rule-based reasoning. The advantage of this strategy is that it eliminates the problem of collecting labeled data in smart environments.
However, the design and implementation of knowledge models require the expertise of domain experts and knowledge engineers, and their capability of handling uncertainty and generalizing is typically limited~\cite{civitarese2019polaris}. 

The recent 
success of Large Language Models (LLMs) paved the way to LLM-based human activity recognition. Preliminary results in the state-of-the-art indicate that LLMs implicitly encode the knowledge about human activities, and hence they can potentially be used to cope with labeled data scarcity~\cite{liu2023large,ji2024hargpt}. Given these promising results, we hypothesize that LLMs could be also leveraged to process data from environmental sensors in smart-home environments to reveal the ADLs performed by the monitored subjects, without the need to collect any labeled data. 
We expect that LLMs would generalize across individuals and environments without requiring any patient- or environment-specific training data, thus offering a scalable approach. Moreover, an LLM-based approach would be more flexible in terms of the set of recognizable ADLs (that is usually fixed in supervised approaches). 

In this work, we propose \acronym{}, a method leveraging pre-trained LLMs to recognize ADLs in smart-home environments. At its core, \acronym{} transforms raw sensor data into natural language sentences describing the high-level events observed in the home. \acronym{} uses a specifically designed prompting strategy to query the LLM about the most likely ADL performed by the subject. Despite \acronym{} does not require training data, we show that a small labeled dataset can further improve the recognition rate thanks to a few-shot prompting mechanism.



The main contributions of this paper are the following:

\begin{itemize}
    \item We propose \acronym{}, a novel LLM-based method for zero-shot sensor-based ADL recognition in smart home environments.
    \item Our results on two public datasets show that zero-shot \acronym{} reaches recognition rates comparable (and sometimes better) than standard supervised approaches.
    \item We extend \acronym{} to perform few-shot prompting when 
    a small labeled dataset is available, by leveraging a semantic-based example selection strategy. Our experiments show that few-shot \acronym{} outperforms standard supervised approaches in labeled data scarcity scenarios.
\end{itemize}

We provide the source code implementing our system to make our results reproducible\footnote{\url{https://github.com/micheleFiori/ADL-LLM}}.

\section{Related Work}
\label{sec:related}

\subsection{Data Scarcity in Human Activity Recognition}


The labeled data scarcity problem in Human Activity Recognition (HAR) is one of the major challenges limiting large-scale real deployments, and several research groups have been working on this problem for several years. Three categories of approaches have been proposed to mitigate labeled data scarcity: data-driven, knowledge-based, and neuro-symbolic.

\subsubsection{Data-Driven Methods}

Most existing approaches to mitigate labeled data scarcity rely on deep learning models. The most common strategies are semi-supervised learning~\cite{abdallah2018activity}, transfer learning~\cite{dhekane2024transfer}, and self-supervised learning~\cite{haresamudram2022assessing}. 
Semi-supervised approaches use small labeled datasets to initialize the model, which is incrementally updated using pseudo-labels inferred on the unlabeled data stream~\cite{abdallah2018activity}. Examples of semi-supervised methods proposed for HAR are self-learning, co-learning, active learning, and label propagation. On the other hand, transfer learning methods leverage models pre-trained on a source domain with a significant amount of labeled data, then adapted in a target domain with labeled data scarcity~\cite{sanabria2021unsupervised, soleimani2021cross}.
Self-supervised learning is a specific transfer learning approach, where the model is pre-trained on massive unlabeled datasets to obtain a reliable feature extractor~\cite{haresamudram2022assessing,jain2022collossl,hiremath2022bootstrapping}. The pre-trained feature extractor is then fine-tuned only using a limited amount of labeled data on the target domain. 
%
%
Existing approaches for self-supervised learning have been proposed for data obtained by wearable and mobile devices, since large datasets (e.g., Biobank~\cite{yuan2024self}, Capture-24~\cite{chan2021capture}) are available in this domain for model pre-training.
To the best of our knowledge, we are not aware of self-supervised learning approaches for smart-home ADL recognition, because in this domain there are only a few small datasets, and learning a comprehensive representation that accommodates the diverse layouts of different homes is still an open challenge.

\acronym{} can be considered a sort of zero-shot transfer learning approach, where a model pre-trained on language is used to recognize activities from a textual representation of sensor data.

\subsubsection{Knowledge-Based Methods}

Several research groups investigated whether common-sense knowledge about human activities could be leveraged to completely avoid labeled data collection. These approaches proposed the use of logic formalisms (e.g., ontologies) to model the relationships between sensor events and activities~\cite{civitarese2019polaris, chen2009ontology}.
However, building such logic models requires skills in the formalism model (e.g., OWL2 in the case of ontologies) and significant domain expertise. This task is usually performed by small groups of persons. Building a comprehensive knowledge model capturing all the possible ways of performing an activity is challenging even for experts, and we are not aware of public ontologies comprehensively covering all the possible human activities and context situations. 

\acronym{} has the same goal of knowledge-based approaches: using common-sense knowledge about human activities to recognize them. However, instead of relying on manually designed logic models, \acronym{} leverages the knowledge implicitly encoded in pre-trained LLM to guarantee better generalization capabilities.

\subsubsection{Neuro-Symbolic Methods}

Neuro-Symbolic AI (NeSy) methods combine the best from data-driven and knowledge-based worlds. Indeed, the goal of NeSy models is to combine symbolic reasoning and machine learning to mitigate data scarcity and improve the model's interpretability~\cite{hitzler2022neuro}.
Considering the HAR domain, logic formalisms have been used to refine the output of the data-driven model~\cite{bettini2020caviar} or to infuse knowledge inside deep learning models~\cite{xing2020neuroplex,arrotta2024semantic}. In these methods knowledge constraints allow the model to capture high-level patterns without learning them from data, thus mitigating labeled data scarcity. However, NeSy methods have been only proposed for wearable-based HAR and their applicability to smart home ADLs recognition has not been explored yet.

For certain aspects, the capabilities of LLMs can be considered as Neuro-Symbolic since they can implicitly mimic reasoning~\cite{fang2024large}. Hence, \acronym{} could be considered as a NeSy model whose data-driven component attempts to mimic symbolic reasoning.

\subsection{Using LLMs for Human Activity Recognition}

In the following, we present relevant research works using LLMs in the HAR domain, distinguishing supervised and zero-shot techniques. Supervised methods combine LLMs and labeled HAR datasets for specific tasks (e.g., prediction, classification), while zero-shot techniques directly utilize pre-trained LLMs to identify activities on unlabeled sensor data.


\subsubsection{LLM-based approaches for supervised activity recognition}

Several research groups studied how to use LLMs for the sensor-based HAR domain.
The most common approach is to fine-tune an LLM with labeled sensor datasets, abstaining an LLM specialized on a specific domain~\cite{liu2023large,okita2023towards}. 
Other works adopted multi-modal models to align sensor data embeddings and text embeddings leveraging contrastive learning to improve the recognition rate~\cite{zhou2023tent}. 
In Neuro-Symbolic HAR, some works used LLMs to infuse common-sense knowledge about human activities into a deep learning model~\cite{arrotta2024contextgpt}. LLMs have also been leveraged to
transform sensor data windows first into sentences in natural language, and then into embeddings to improve the recognition rate and generalization capabilities of supervised models~\cite{thukral2024layout}.
Finally, since LLMs are pre-trained to predict the next word in a sentence, some research groups fine-tuned them also to predict future sensor events~\cite{takeda2023sensor,graule2023gg}. 


\subsubsection{Zero-shot approaches for human activity recognition using LLMs}

Similarly to \acronym{}, other works aim at recognizing human activities using LLMs without any training phase. 
For instance, some works show that LLMs can reason on raw inertial sensor data from mobile/wearable devices to distinguish between a limited set of low-level physical activities~\cite{ji2024hargpt, hota2024evaluating}. 
Another line of research leverages a pre-trained model generating 3D motion sequences from text~\cite{zhang2023t2m} to generate synthetic labeled data from wearable/mobile devices~\cite{leng2024imugpt}.
Considering smart environments, ChatGPT has been used for zero-shot activity recognition~\cite{xia2023unsupervised}. However, this work assumes the availability of several sensors capturing the interaction of the subject with many household items (e.g., specific food ingredients), which is unrealistic in real-world deployments. In this work, we consider a more realistic setting considering the sensors that are usually deployed in smart homes for ADL recognition (e.g., motion sensors, magnetic sensors, plug sensors).
Finally, another work uses an LLM to provide a label to each cluster of raw sensor data time windows (collected in an initial deployment phase)~\cite{gao2024unsupervised}. A supervised classifier is then trained using the pseudo-labels obtained through the LLM. 
However, such an approach is susceptible to the quality of clustering and, once the supervised classifier is trained, it may fail in capturing behavioral changes or activities not performed while collecting unlabeled data. Differently from that work, \acronym{}~aims at leveraging the flexibility of LLMs for each observed time window.

\section{The \acronym{} framework}

In this section, we present \acronym{}: our LLM-based method for ADL recognition in smart home environments.

\subsection{Problem Formulation}
We consider a classic problem formulation for ADL recognition. 
A smart home is equipped with a set $\mathbf{S} = \{S_1,S_2,\dots,S_m\}$ of environmental sensors. Examples are magnetic sensors revealing open-close events (e.g., doors, cabinets, etc.), motion/presence sensors revealing movement/presence in specific areas, and smart plugs revealing the use of appliances. 
There is a set of target activities of interest $\mathbf{A} = \{A_1, A_2,\dots, A_n\}$, usually suggested by clinicians.
%
As in most works in this domain, we consider a smart home environment inhabited by a single subject or with a sensing infrastructure able to properly associate a sensor's activation with the subject that activated it. 
Based on the interaction of the subject with the environment, the smart home generates a continuous stream of sensor activations and de-activations. This stream is partitioned in fixed-time windows of $\tau$ seconds with a given overlapping factor $o$.
Given a window $w_t$, the goal of \acronym{} is to determine the activity $A \in \mathbf{A}$ that the subject was more likely performing in the interval $[t,t+\tau]$.

\subsection{Overall architecture of \acronym{}}

\begin{figure}[h!]
    \centering
    \includegraphics[width=0.8\textwidth]{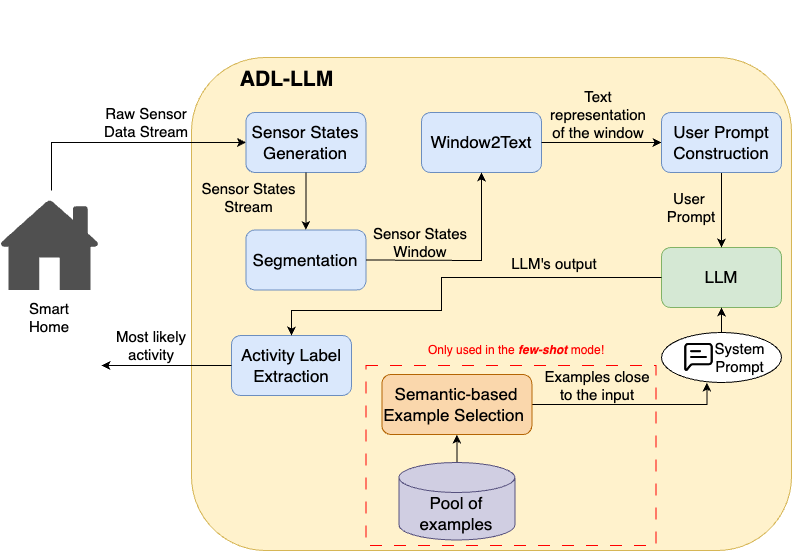}
    \caption{Overall architecture of \acronym{}. When the pool of examples is empty \acronym{}, acts as a \textbf{\textit{zero-shot}} ADLs recognition method. Otherwise, it is a \textbf{\textit{few-shot}} approach.}
    \label{fig:adl-llm}
\end{figure}

We assume that \acronym{} runs on a trusted platform.
In the case of the medical application, it may run on a telemedicine platform
as one of the components of a complex system that continuously monitors subjects at risk of cognitive decline. 
Figure~\ref{fig:adl-llm} shows the overall architecture of the \acronym{} framework.

The smart home generates a continuous stream of raw sensor data that is pre-processed by the \textsc{Sensor State Generation} module to obtain a stream of sensor states. Intuitively, a sensor state represents the timespan in which a sensor is \textit{active} (e.g., ``\textit{the TV in the living room is ON from 12:34 to 13:02'}'). Then, the stream of sensor states is segmented in fixed-time windows by the \textsc{Segmentation} module. 

Since LLMs are specialized in natural language text, the main idea of \acronym{} is to generate a natural language representation for each window, so that an LLM can ``reason'' on it leveraging its implicit common-sense knowledge. 
The \textsc{Window2Text} module is in charge of transforming windows of sensor states into natural language sentences describing what happened in the home during that window.
The LLM receives as input a \textsc{System Prompt} (i.e., instructions about the task it has to perform) and a \textsc{User Prompt} (i.e., the textual representation of the window enriched with additional instructions for the LLM by the \textsc{User Prompt Construction} module). The output of the LLM is finally processed by the \textsc{Activity Label Extraction} module to obtain the most likely activity performed by the subject.

\acronym{} works in two different modalities:

\begin{itemize}
    \item \textbf{\textit{Zero-Shot}}, where no labeled data is available
    \item \textbf{\textit{Few-Shot}}, where a small labeled dataset is available
\end{itemize}

Indeed, in certain circumstances, it may be possible to collect a small dataset of sensor data windows labeled with the corresponding ADL. For instance, when the sensing infrastructure is physically installed in the home, the technicians may ask the subject to perform a limited number of activities (as a sort of "system calibration" phase). The data associated with these activities is manually labeled and stored in the \textit{Pool of Examples}. In few-shot \acronym{}, when processing a window, the \textsc{System Prompt} is enriched by the \textsc{Semantic-based Example Selection} module, which picks from this pool the samples that are the most ``semantically'' similar to the textual representation of currently processed window. The selected samples have the role of further guiding the LLM on the ADLs recognition task. In the LLM domain, this approach is known as \textit{few-shot prompting}, and it is widely used to improve the output of LLMs~\cite{brown2020language}.






\subsection{The main components of \acronym{}}

In the following, we describe the main modules of \acronym{}, i.e., those used both in the zero-shot and the few-shot modalities.

\vspace{2mm}
\subsubsection{Sensor States Generation}

The smart home generates a continuous stream of binary sensor events $Stream = \langle \dots, E_{t-1}^{S_l}, E_t^{S_m}, E_{t+1}^{S_q}, \dots \rangle$. Each $E_t^S$ represents the activation or the deactivation of a sensor $S$ at a specific time $t$. For instance, consider a magnetic sensor on the fridge. This sensor generates an ON event when activated (when the subject opens the fridge) and an OFF event when deactivated (when the subject closes the fridge). Other sensors share a similar behavior, like motion and pressure sensors. For sensors that generate continuous values, as the smart plug that continuously detects the power consumption of specific appliances, we use a threshold on the detected energy consumption to infer when the appliance is turned ON and OFF. 
For sensors that provide almost continuous values but are not directly associated with activation and deactivation events, like the temperature sensor, we discretize the values in different relevant ranges (e.g., for the temperature: $15$-$18$\textit{°C}, $19$-$24$\textit{°C}, $24$-$30$\textit{°C}, $>$$30$\textit{°C}). Suppose that the thermostat in the living room detected $18$°C at $t_i$ and then $19$\textit{°C} at time $t_{i+1}$. In this case, the stream includes a deactivation event at time $t_i$ for the temperature being in the range $15$-$18$\textit{°C}, and an activation event at time $t_{i+1}$ for the temperature being in the range $19$-$24$\textit{°C}. 

We convert the stream of events into a stream of sensor states, where a sensor state $st(\sigma_i, t_s, t_e)$ denotes that the sensor state $\sigma_i$ (derived from $S_i$) is true during the time interval $[t_s,t_e]$. For instance, $st(\textit{FridgeDoorOpen}, \textit{12:32pm},\textit{12:33pm})$ represents the fact that the fridge door was open in the interval $[\textit{12:32pm},\textit{12:33pm}]$. As another example, the $st(\textit{NearTheStove}, \textit{11:45am},\textit{12:00pm})$ represent the fact that the motion sensor near the stove detected the subject's movements from \textit{11:45am} to \textit{12:00pm}.

More formally, we generate a sensor state $st(\sigma,t_s,t_e)$ when the stream contains two events $E_{t_s}^S$ and $E_{t_e}^S$ with $t_s < t_e$ being the instants in which the events of activation and de-activation of sensor $S$ occurred, $\sigma$ indicates a property revealed by the activation state of $S$, and there is no $E_{t_q}^S$ such that $t_s<t_q<t_e$.

\vspace{2mm}
\subsubsection{Segmentation}

\acronym{} employs a fixed-time segmentation of the stream of sensor states, considering windows of $\tau$ seconds and an overlap factor $o$. Intuitively, each window is associated with all the sensor states whose interval intersects with the window interval. More formally, a window 
$w_t$ over the interval $[t,t+\tau]$ is associated with the sensor states $st(\sigma,t_s,t_e)$ such that $t_s \leq t+\tau$ and $t_e \geq t$. 

Note that a window may be associated with multiple states even derived from different sensors. Based on the time intervals, we distinguish three categories that will be useful for text generation:

\begin{itemize}
    \item \textit{Inner states}: this is the case of $t_s \geq t$ and $t_e \leq t+\tau$. This represents the simple case of the state occurring completely within the window. 
    \item \textit{Already active states}: this is the case of $t_s < t$. These states are already active before the start of the window. 
    \item \textit{Persistent states}: this is the case of $t_e>t+\tau$. These states persist after the end of the window. 
\end{itemize}

Note that a state may be both \textit{Already active} and \textit{Persistent} when $t_s<t$ and 
$t_e>t+\tau$.

\vspace{2mm}
\subsubsection{Window2Text}
\label{subsec:w2t}


The \textsc{Window2Text} module converts a window $w_t$ into its corresponding textual representation $T(w_t)$. 
In the following, we provide a high-level description of how this conversion is done:
\begin{itemize}
    \item Each generated  $T(w_t)$ begins by reporting the window's length ($\tau$ seconds) and the approximate time of the day (in the format hh:mm am/pm). The first was found useful to provide a temporal context to the LLM since $\tau$ is usually shorter than the typical duration of an ADL; The second was found useful for the LLM to exploit correlations between common ADLs and the time of the day in which they are typically executed (e.g., \textit{preparing dinner}, \textit{preparing lunch}).  %
    \item After reporting the temporal context, $T(w_t)$ describes what happened in the home during $w_t$ from the subject's perspective (e.g., "the subject opened the fridge"). This choice, rather than considering a sensor's perspective as done in other works~\cite{thukral2024layout} (e.g., "the magnetic sensor on the fridge was activated"), leverages the LLM implicit knowledge of how humans perform activities. Indeed, subject-based descriptions are likely more common in LLM training data, and hence they may lead to more accurate results than indirectly interpreting actions through sensor events. 
    \item The subject's perspective description of a window $w_t$ has the following structure:
    \begin{itemize}
    \item First, it reports the location where the subject is at the beginning of the window (if this information can be inferred from sensor data).
    \item Then, it reports the \textit{Already active states} (e.g., ``the subject is already sitting on the couch'').
    \item Finally, for each room visited by the subject during $w_t$ (in chronological order), the description includes the actions (derived by the states) performed by the subject during $w_t$ generated leveraging the \textit{Inner} and \textit{Persistent} states.
    \end{itemize}
    \item To transform each state into actions described in natural language, we implemented simple rules leveraging its category, its property $\sigma$, its relative temporal order with the other states in $w_t$, its duration, and the semantic location of the corresponding sensor. Hence, \textsc{Window2Text} assumes the availability of meta-data describing how to map each sensor $S$ to its location in the home and the corresponding $\sigma$.
\end{itemize}

Since LLMs often struggle in reasoning with numbers and mathematics, in \textsc{Window2Text} we preferred to avoid absolute timestamps, considering instead relative temporal relationships between states by adopting temporal adverbs (e.g., \textit{``Then''}, \textit{``After $3$ seconds''}) and reporting the duration of the states (e.g., \textit{``for $5$ seconds''}). Since we provide an anchor to absolute time by reporting the approximate time of the day at the beginning of the window description, 
the reported events are also approximately anchored in time.
The above strategy is based on the observation that LLMs tend to perform better with qualitative information, as they can focus on understanding the sequence and relationships of sensor states rather than managing exact time calculations.
In Section~\ref{subsec:alternativeprompts}, we compare our approach with an alternative prompt using exact timestamps.    

\begin{figure*}[t]
    \centering
    \includegraphics[width=0.8\linewidth]{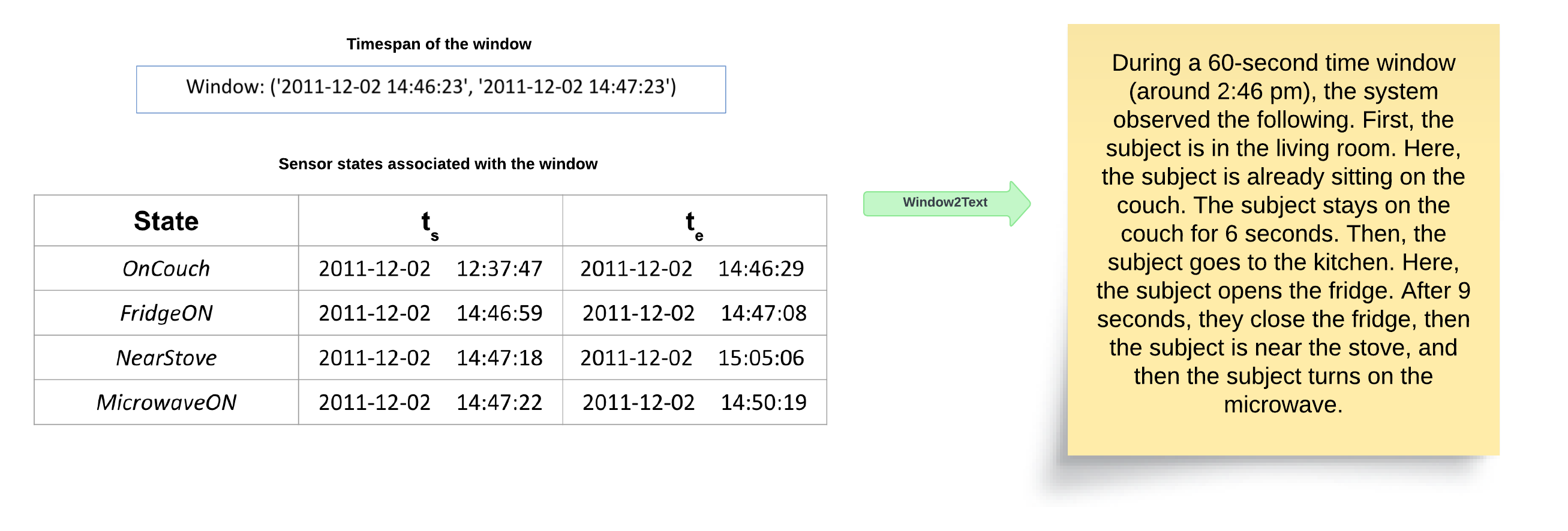}
    \caption{An example of \textsc{Window2Text} in action on the UCI ADL dataset}
    \label{fig:w2t}
\end{figure*}

Figure~\ref{fig:w2t} shows an example of \textsc{Window2Text} 
application
on the UCI ADL dataset.

In the following, we show examples of how \textsc{Window2Text} 
generates the subjects' perspective description of what happened in the home based on sensor states. In particular, Example~\ref{ex:time} shows how the temporal nature of each state affects the generated sentence, while Example~\ref{ex:rooms} shows a more complete example also taking into account the location information.

\begin{example}
\label{ex:time}
If the state 
$st(StoveON,t_s,t_e)$ 
is an inner state in the current window, it is mapped to the sentence ``The subject turned on the stove and turned it off after $t_e-t_s$ seconds''. When the same state is an \textit{Already active} state for the current window, the sentence ``The stove is already turned on'' is produced, and if there is another already active state (e.g., about the fridge) the corresponding sentence is combined with the one for the stove. When the $StoveON$ state ends before the end of the window, the following sentence is produced ``After $t_e-t$ seconds the subject turned off the stove".
As another example, the inner state $st(NearBathroomSink,t_s,t_e)$ generated by a PIR sensor leads to the sentence ``The subject was near the sink for $t_e-t_s$ seconds''.
\end{example}
\mbox{}\\ 
\\
\begin{example}
\label{ex:rooms}
Consider a window $w_t$ that has associated the \textit{Inner states} $st(FridgeDoorOpen,t1,t2)$, $st(NearStove,t3,t4)$ and a \textit{Persistent state} $st(OnTheCouch,t5,t6)$, where $t2<t3$, $t4<t5$, and $t6>t+\tau$. \acronym{} automatically infers (using simple  rules) that $FridgeDoor$ and $NearStove$ occurred consecutively in \textit{kitchen}, and after the subject moves to the \textit{living room} for the $OnTheCouch$ state. This window leads to the following natural language sentence: ``The subject is in the kitchen. Here, they open the fridge door. After $t2-t1$ seconds, they close the fridge door. Then, they are near the stove for $t4-t3$ seconds. Then, they move to the living room. Here, they sit on the couch.''
\end{example}
\mbox{}\\

\vspace{2mm}
\subsubsection{System prompt}
\label{subsec:sysprompt}

\begin{figure}[h!]
    \centering
    \includegraphics[width=0.6\textwidth]{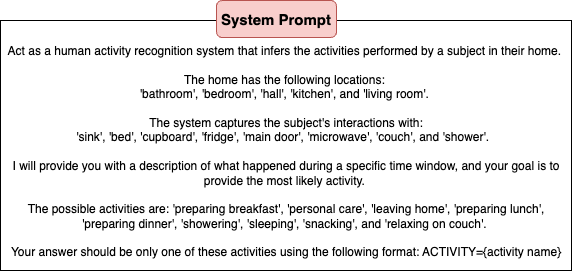}
    \caption{An example of system prompt in \acronym{} for the UCI ADL dataset (Home B). The sets of locations, household elements captured by sensors, and activities are variables depending on the specific home where \acronym{} is installed.}
    \label{fig:zero-prompt}
\end{figure}

The system prompt is the set of instructions defining the general task assigned to the LLM and is independent of the specific window being considered. Our prompt first asks the LLM to act as an activity recognition system. Then, it provides information about the rooms in the smart home, the main interactions with the environment that can be captured by the available sensors, and the  list of activities of interest.
Since this information depends on the specific smart environment and application, it must be provided during the installation phase.
An example of a system prompt is reported in Figure~\ref{fig:zero-prompt}.

\vspace{2mm}
\subsubsection{User Prompt Construction}
The user prompt contains the information about the specific window $w_t$ for which the LLM has to predict the most likely activity. 
The user prompt contains:
\begin{itemize}
    \item The textual representation $T(w_t)$ of the window obtained with \textsc{Window2Text} (described in Section~\ref{subsec:w2t}). 
    \item Instructions about selecting only one activity among the target ones. 
    \item The sentence ``Reason step by step'', used to instruct the LLM to perform intermediate ``reasoning'' steps before providing the predicted ADL. This technique is well-known as Chain Of Thought (COT), it usually leads to increasing the correctness of the LLM's output~\cite{wei2022chain}, and proved to be useful even in our system. 
\end{itemize}

\vspace{2mm}
\subsubsection{LLM-based ADLs Recognition and Activity Label Extraction}

\begin{figure}[h!]
    \centering
    \includegraphics[width=0.55\textwidth]{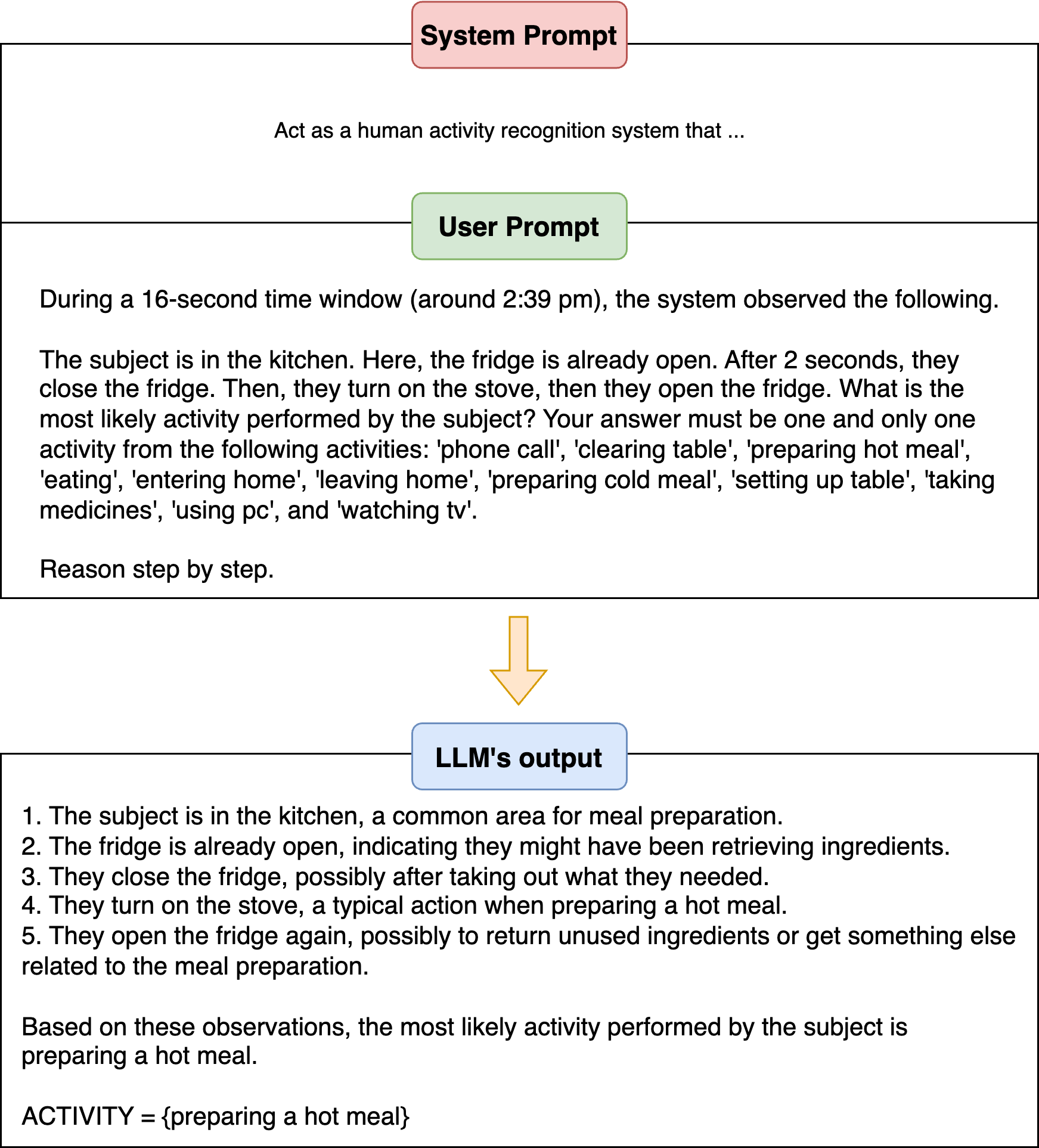}
    \caption{\acronym{}: An example of input and corresponding output.}
    \label{fig:output}
\end{figure}

The system prompt and the user prompt are provided as input to the LLM for each window $w_t$, and the LLM provides a textual description as output, reporting some facts supporting the selection of the predicted activity. 
%
Figure~\ref{fig:output} shows an example of the output returned by the LLM.

Finally, the \textsc{Activity Label Extraction} module uses regular expressions to extract only the activity label from the LLM's output. In rare cases, despite the instructions, the LLM provides an activity label that does not syntactically match any of the candidate activities in $A$. For instance, instead of \textit{snacking}, in a few cases, the LLM returned as output \textit{preparing late night snack} or \textit{preparing midnight snack}. In these cases, the \textsc{Activity Label Extraction} module uses the cosine similarity to compare the text embeddings of the labels. The label with the highest similarity to the predicted output is selected as the final prediction.



\subsection{Few-shot \acronym{} components}
When a small dataset of labeled sensor data is available, \acronym{} can be improved by implementing few-shot prompting. This is achieved by extending the zero-shot architecture with two components described below.

\vspace{2mm}
\subsubsection{Building a Pool of Examples}
\label{subsec:poolofexamples}

%
In some cases, obtaining a limited amount of sensor data with the associated ground truth (ADLs) may be possible. For example, a technician may perform annotation during an initial calibration phase of the system, or some limited dataset collected in similar settings as the deployment target may become available.
Let $\mathbf{D}$ be this dataset composed of a set of pairs $(w, A)$ where $w$ is a window of raw sensor data, while $A$ is the corresponding activity label. 
We convert $\mathbf{D}$ to a set $\mathbf{E}$ called \textit{Pool of Examples}. This set contains the textual representation of the windows in $\mathbf{D}$ (leveraging \textsc{Window2Text}) associated with the corresponding activity label. More formally, for each pair $(w,A)\in\mathbf{D}$, $\mathbf{E}$ includes a pair $(T(w),A)$.  Note that, since $\mathbf{E}$ is a set, there is only one occurrence for each pair $(T(w),A)$. We also remove from $\mathbf{E}$ all the examples sharing the same textual description $T(w)$ but associated with different labels. Indeed, such examples represent patterns shared by multiple activities, and considering these as additional information may be misleading for the LLM.


\vspace{2mm}
\subsubsection{Semantic-based Example Selection}

\begin{figure}[h!]
    \centering
    \includegraphics[width=0.5\textwidth]{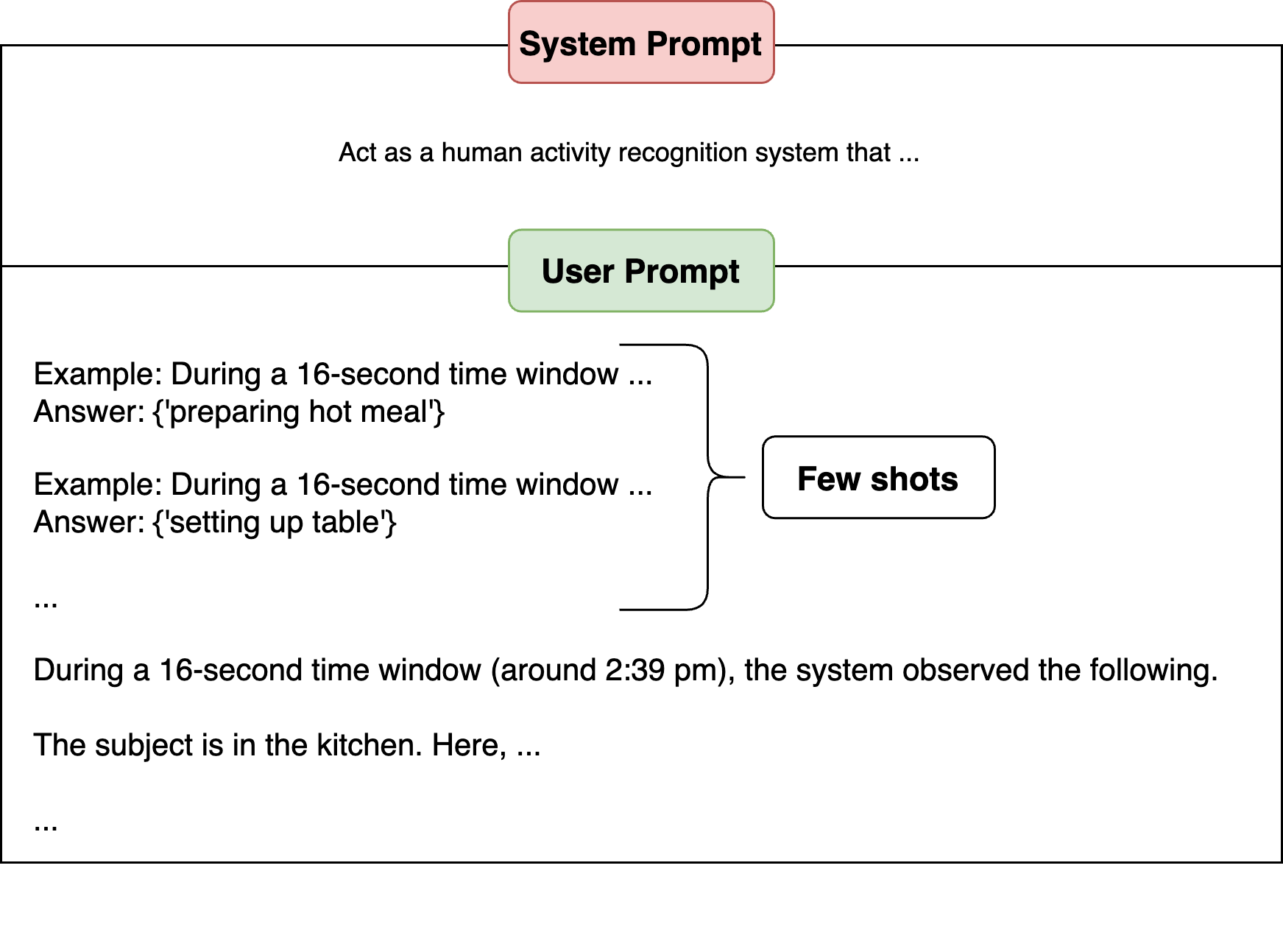}
    \caption{\acronym{}: An example of the prompt in the few shot setting.}
    \label{fig:few-shot-prompt}
\end{figure}

In principle, we may
include the whole set $\mathbf{E}$ in each user prompt as additional context for~\acronym{}. However, this would introduce excessive costs (both in terms of money and performance), since each prompt would include a long sequence of textual examples. Hence, following the Retrieval Augmented Generation (RAG) 
principle~\cite{lewis2020retrieval}, we include only the $k$ examples in $\mathbf{E}$ that are the most semantically similar to the input window.
More formally, given a textual representation $T(w_t)$ of a window $w_t$ currently processed by \acronym{}, we compute the corresponding embedding $\phi_{w_t} = \Phi(T(w_t))$,
where $\Phi$ is a function (e.g., a pre-trained LLM) mapping sentences into a latent space representation (i.e., their embedding).
Then, for each pair $(T(w),A) \in \mathbf{E}$, we compute the pairwise cosine similarity between $\phi_{w_t}$ and $\phi_w = \Phi(T(w))$.
We enrich the system prompt with the $k$ pairs $(T(w), A) \in E$ that are associated with the highest pairwise similarity between $\phi_{w_t}$ and $\phi_w$. The pseudo-code for \textsc{Semantic-Based Example Selection} is summarized in Algorithm~\ref{alg:fewhost}. Figure~\ref{fig:few-shot-prompt} shows an example of user prompt of Few-Shot~\acronym{}.

\renewcommand{\algorithmicrequire}{\textbf{Input:}}
\renewcommand{\algorithmicensure}{\textbf{Output:}}

\begin{algorithm}[h!]
\footnotesize
\caption{Semantic-based Example Selection}
\label{alg:fewhost}
\begin{algorithmic}
    \REQUIRE The pool of examples $\mathbf{E}$, the textual representation of the current window $T(w_t)$, an embedding function $\Phi$, the number of examples to be retrieved $k \in \mathbb{N^+}$
    \ENSURE the $k$ examples from $\mathbf{E}$ that are the most semantically similar to $T(w_t)$
    \STATE $\phi_{w_t} \gets \Phi(T(w_t))$ 
    \STATE $pairwiseDict \gets \{\}$ \textit{// Initialize an empty dictionary}
    \FOR{$(T(w),A) \in \mathbf{E}$}
        \STATE $\phi_w \gets \Phi(T(w))$ 
        \STATE $pairwiseDict[T(w)] \gets \textit{cosineSimilarity}(\phi_{w_t},\phi_w)$
    \ENDFOR
    \RETURN  $\textit{top } k \textit{ examples in } pairwiseDict$
\end{algorithmic}
\end{algorithm}

\section{Experimental Evaluation}
\label{sec:experiments}

\subsection{Datasets}
\label{subsec:datasets}

We evaluated \acronym{} using two public datasets: MARBLE~\cite{arrotta2021marble} and UCI ADL~\cite{ordonez2013activity}. We chose these datasets because they include sensor data from a wide variety of sensors (e.g., magnetic, plug, motion) that we are currently considering on a project involving data collection in the homes of subjects with Mild Cognitive Impairment (MCI). While the majority of works in the literature utilize the well-known CASAS datasets~\cite{cook2012casas}, which primarily include data from motion sensors, these datasets are limited in their ability to precisely characterize the specific activities performed by subjects. Although data-driven models can identify low-level patterns within motion sensor data, our goal is to leverage the common-sense knowledge implicitly encoded in LLMs. Preliminary experiments suggested that using LLMs with datasets mostly based on motion sensors can not reliably recognize ADLs. 


\subsubsection{MARBLE}

The MARBLE dataset~\cite{arrotta2021marble} includes data collected from $12$ subjects in both single and multi-inhabitant scenarios. 
The dataset was collected using the following environmental sensors: magnetic sensors to capture the opening and closing of various drawers and cabinets, pressure sensors to detect when a subject was sitting at the dining table or on the sofa in the living room, and smart plugs to monitor the usage of the stove and television. Additionally, the dataset includes data from an Android application on the user's smartphone, which automatically detects incoming and outgoing phone calls. Finally, the dataset incorporates data from a smartwatch worn by the user, providing micro-localization information (based on an infrastructure using Bluetooth Low Energy (BLE) beacons) at the room level.

We performed our experiments only considering the single-inhabitant scenarios.
Moreover, while the dataset also included inertial sensor data from the smartwatch (e.g., continuous acceleration on three axes), we ignore this type of data as it can not be directly transformed into a prompt in natural language.

The dataset includes the following $13$ activities: \textit{answering phone}, \textit{clearing table}, \textit{preparing hot meal}, \textit{preparing cold meal}, \textit{eating}, \textit{entering home}, \textit{leaving home}, \textit{making phone call}, \textit{setting up table}, \textit{taking medicines}, \textit{using PC}, \textit{washing dishes}, and \textit{watching TV}. 

In our experiments, we do not consider the \textit{washing dishes} activity since it is mainly characterized by inertial sensors in smartwatch data (not used by \acronym{}). Moreover, we merged the \textit{making phone call} and \textit{answering phone call} activities into a single activity called \textit{phone call}.

\subsubsection{UCI ADL}

The UCI ADL dataset~\cite{ordonez2013activity} includes data about the ADLs performed by two subjects in their own homes. The two homes are named \textit{Home A} and \textit{Home B} in the dataset. The overall data collection duration is $35$ days: $14$ days for Home A and $21$ days for Home B.

Different environmental sensors are considered: PIR sensors (installed on doors, bathroom sink, stove, and shower areas) to detect the subject's movements, pressure sensors to detect when the subject was sitting on the couch and the bed, magnetic sensors to detect when the subject was opening and closing cupboards or cabinets, and plug sensors to detect the usage of microwave and toaster. 

Home A and Home B sensing infrastructure have some differences. Compared to Home A, Home B has a slightly more limited sensing infrastructure since it is not equipped with the PIR sensor over the stove, the plug sensor to detect the usage of the toaster, and the magnetic sensor to detect interactions with the bathroom cabinet. Nonetheless, Home B differently from Home A,  includes additional PIR sensors installed on the kitchen, bedroom, and living room doors.

The dataset covers the following activities: \textit{preparing breakfast}, \textit{preparing lunch}, \textit{preparing dinner}, \textit{snacking}, \textit{personal care} (labeled as grooming in the dataset), \textit{showering}, \textit{leaving},\textit{relaxing on couch} (labeled as \textit{spare time/TV} in the dataset), \textit{sleeping}, and \textit{toileting}. Note that \textit{preparing dinner} occurs only in the dataset collected in Home B.
In our experiments, we ignored the \textit{toileting} activity since the real semantics of this activity and of the flush sensor were not clear in the dataset (and it was not reported in the dataset's documentation).

\subsection{Supervised baselines}

To show the effectiveness of \acronym{}, we compare it with three state-of-the-art supervised approaches for ADL recognition.

\begin{itemize}
    \item \textbf{MLP}~\cite{diallo2021human}: As the simplest baseline, we adopted a Multi-Layer Perceptron (MLP) architecture that was proposed for smart-home HAR.
    \item \textbf{BiLSTM}\cite{liciotti2020sequential}: Also known as \textit{Bidirectional LSTM}. It represents a standard recurrent neural network based on LSTM, processing windows in both forward and backward directions.
    \item \textbf{DeepConvLSTM}~\cite{ordonez2016deep}: Also known as \textit{Deep Convolutional and LSTM Recurrent Neural Network}. It combines convolutional and LSTM layers to effectively capture both spatial and temporal features.
\end{itemize}

These baselines are representative because they reflect the most effective architectures used in the literature for supervised ADL recognition. We did not include methods specifically designed for labeled data scarcity scenarios because we are not aware of state-of-the-art approaches specifically designed for this setting.
As we already mentioned in Section~\ref{sec:related}, methods like self-supervised learning or neuro-symbolic approaches have been proposed only for mobile/wearable-based HAR. Smart-Home ADL recognition introduces different challenges, since there are no large datasets available for pre-training, and different homes may have significantly different sensor layouts.

\subsection{Experimental Setup}

\subsubsection{Implementation}
We implemented a prototype of \acronym{} in Python 3.10.12 language using the langchain 0.1.17 library.
As LLM for ADLs recognition, we used the OpenAI \textit{`gpt-3.5-turbo-0125'} model (setting temperature to $0$), while we used the OpenAI \textit{`text-embedding-3-small'} model to compute embeddings from the textual representation of windows for few-shot prompting. Based on what is suggested in the literature, we used $16$-second windows with $80\%$ overlap for the MARBLE dataset~\cite{arrotta2021marble} and $60$ seconds windows with $80\%$ overlap for the UCI ADL dataset~\cite{ordonez2013activity}. 

Considering the supervised baselines, we used the code available in public repositories (DeepCASAS~\footnote{https://github.com/danielelic/deep-casas/} and DeepConvLSTM~\footnote{https://github.com/STRCWearlab/DeepConvLSTM/}). We slightly modified the existing implementations to ensure compatibility with our software environment and data structures, while faithfully reproducing the architectures described in the literature. In all of our baselines, the LSTM layers have $64$ units, while for the convolutional block of DeepConvLSTM we used a single convolutional layer with $8$ $2$x$2$ filters. Before feeding it to the networks, environmental sensor data is converted into multivariate binary time-series following a standard approach in the literature~\cite{van2008accurate}.


\subsubsection{Dataset splitting strategy}
Since our major contribution is a zero-shot approach, in principle we could have used the entire dataset as a test set.
However, we also need a training set 
for training the supervised baselines and as an example dataset for the few-shot \acronym{}.
For this reason, we divided each dataset into $30\%$ for the training set and $70\%$ for the testing set. This splitting reflects a labeled data scarcity scenario, where only a limited labeled dataset is available. The training set is not used in any way by zero-shot \acronym{}, while it is used to train the supervised baselines and to generate the pool of examples in few-shot \acronym{}.

In our experiments,  the training and test sets contain data from the same user. This is partly because some of the considered datasets include only one user and partly because in real deployments the small Pool of Examples used by few-shot~\acronym{} may be collected from the target home during the system's initial deployment.
Nevertheless, no window of sensor data in the training set appears in the test set, ensuring that the model is not exposed to data leakage. 
Note that the supervised baselines would likely exhibit even lower performance using a user-based split; hence the advantage of zero-shot~\acronym{} would be even clearer. 


\subsubsection{Simulating data scarcity scenarios}

In order to evaluate the effectiveness of few-shot \acronym{} we consider different labeled data availability scenarios to populate the pool: $100\%$, $50\%$, $25\%$, $10\%$, and $5\%$ of the training set. In these settings, the supervised baselines use the available labeled data for training. We expect that, in data scarcity scenarios, few-shot \acronym{} outperforms the supervised baselines.

\subsection{Results}

\subsubsection{Zero-shot \acronym{}}\mbox{}\\

Tables~\ref{tab:zs-uci} and~\ref{tab:zs-marble} show, for both datasets, the recognition rate of zero-shot \acronym{} (not using any labeled data) with the supervised baselines trained using the $100\%$ of the training set.

\begin{table}[h!]
\tiny
\caption{UCI ADL: F1 score of Zero-Shot (using no training data) \acronym{} vs. supervised baselines trained using $100\%$ of the training set (i.e., $30\%$ of the dataset)}
\label{tab:zs-uci}
\centering
\begin{tabular}{|r|c|c|c|c|c|c|c|c|}
\hline
& \multicolumn{4}{c|}{ Home A } & \multicolumn{4}{c|}{ Home B} \\ \hline
\multicolumn{1}{|c|}{Activity} & \acronym{} & MLP & BiLSTM & DeepConvLSTM & \acronym{} & MLP & BiLSTM & DeepConvLSTM \\  \hline 
Leaving home        &  1.00 & 0.97 & 0.95  & 0.99 & 0.90 & 0.85 & 0.84 & 0.88 \\ \hline
Personal care       &  0.97 & 0.99 & 0.98  & 0.99 & 0.96 & 0.94 & 0.92 & 0.93 \\ \hline
Preparing breakfast &  0.75 & 0.92 & 0.90  & 0.94 & 0.71 & 0.29 & 0.35 & 0.39 \\ \hline
Preparing dinner    &   --  & --         & --   & --   & 0.33 & 0.16 & 0.06 & 0.13 \\ \hline
Preparing lunch     &  0.91 & 0.96 & 0.95  & 0.96 & 0.40 & 0.30 & 0.35 & 0.41 \\ \hline
Relaxing on couch   &  0.96 & 1.00 & 1.00  & 1.00 & 0.71 & 0.86 & 0.85 & 0.87 \\ \hline
Showering           &  0.99 & 0.99 & 0.99  & 1.00 & 0.95 & 0.95 & 0.93 & 0.96 \\ \hline
Sleeping            &  1.00 & 1.00 & 0.99  & 1.00 & 0.94 & 0.99 & 0.99 & 0.99 \\ \hline
Snacking            &  0.36 & 0.00 & 0.04  & 0.00 & 0.28 & 0.39 & 0.37 & 0.40 \\ \hline
\hline
\textbf{Weighted Avg. } &  0.94 & 0.97 & 0.96 & 0.97 & 0.75 & 0.73 & 0.72 & 0.75 \\ \hline
\end{tabular}
\end{table}

\begin{table}[h!]
\tiny
\caption{MARBLE: F1 score of Zero-Shot \acronym{} (using no training data) vs. supervised baselines using $100\%$ of the training set (i.e., $30\%$ of the dataset)}
\label{tab:zs-marble}
\centering
\begin{tabular}{|l|c|c|c|c|}
\hline
Activity            & \acronym{} &  MLP  & BiLSTM & DeepConvLSTM \\ \hline
Clearing table      & 0.25       & 0.40 & 0.44 & 0.55 \\ \hline
Eating              & 0.91       & 0.91 & 0.91 & 0.91 \\ \hline
Entering home       & 0.41       & 0.00 & 0.00 & 0.13 \\ \hline
Leaving home        & 0.71       & 0.00 & 0.00 & 0.00 \\ \hline
Phone call          & 0.97       & 0.80 & 0.76 & 0.81 \\ \hline
Preparing cold meal & 0.51       & 0.70 & 0.63 & 0.73 \\ \hline
Preparing hot meal  & 0.82       & 0.89 & 0.82 & 0.90 \\ \hline
Setting up table    & 0.13       & 0.58 & 0.55 & 0.57 \\ \hline
Taking medicines    & 0.47       & 0.57 & 0.54 & 0.59 \\ \hline
Using pc            & 0.97       & 0.91 & 0.91 & 0.91 \\ \hline
Watching TV         & 0.97       & 0.95 & 0.94 & 0.96 \\ \hline
\hline
\textbf{Weighted F1-score } & 0.80 & 0.81 & 0.78 & 0.82 \\ \hline
\end{tabular}
\end{table}

We observe that, for all the datasets, zero-shot \acronym{} reaches recognition rates that are very close to those of DeepConvLSTM, which is the best among the supervised baselines. Indeed, on the UCI ADL dataset, \acronym{} is only $3\%$ behind DeepConvLSTM in terms of F1 score, while only $2\%$ on the MARBLE dataset. 
We observe that Home A of the UCI ADL dataset represents the easiest benchmark while Home B, having a reduced sensing infrastructure (like we explained in Section~\ref{subsec:datasets}), makes ADL recognition more challenging. Interestingly, we observe that the MLP baseline is the second-best and achieves results close to those of DeepConvLSTM. This suggests that, in this domain, the input data is relatively simple and does not require complex automatic feature extraction, allowing the MLP to perform competitively despite its simpler architecture.

An interesting insight is that sometimes \acronym{} is capable of recognizing ADLs that are not captured by supervised baselines. For instance, this happens in the Home A of the UCI ADL dataset, where \acronym{} is capable of recognizing \textit{snacking} (even if with a low F1 score), while the supervised baselines fail. Similar phenomena occur in Home B for \textit{preparing dinner} and in MARBLE for \textit{entering home} and \textit{leaving home}. This is because these activities are poorly represented in their respective datasets, and a supervised classifier can not learn them properly during the training phase (considering that the training set is $30\%$ of the whole dataset). On the other hand, \acronym{} leverages the common-sense knowledge implicitly encoded in the LLM, and it is capable of recognizing these activities without any type of training.

\begin{figure}[h!]
  \centering
  \begin{subfigure}[b]{0.49\linewidth}
    \centering
    \includegraphics[width=\linewidth]{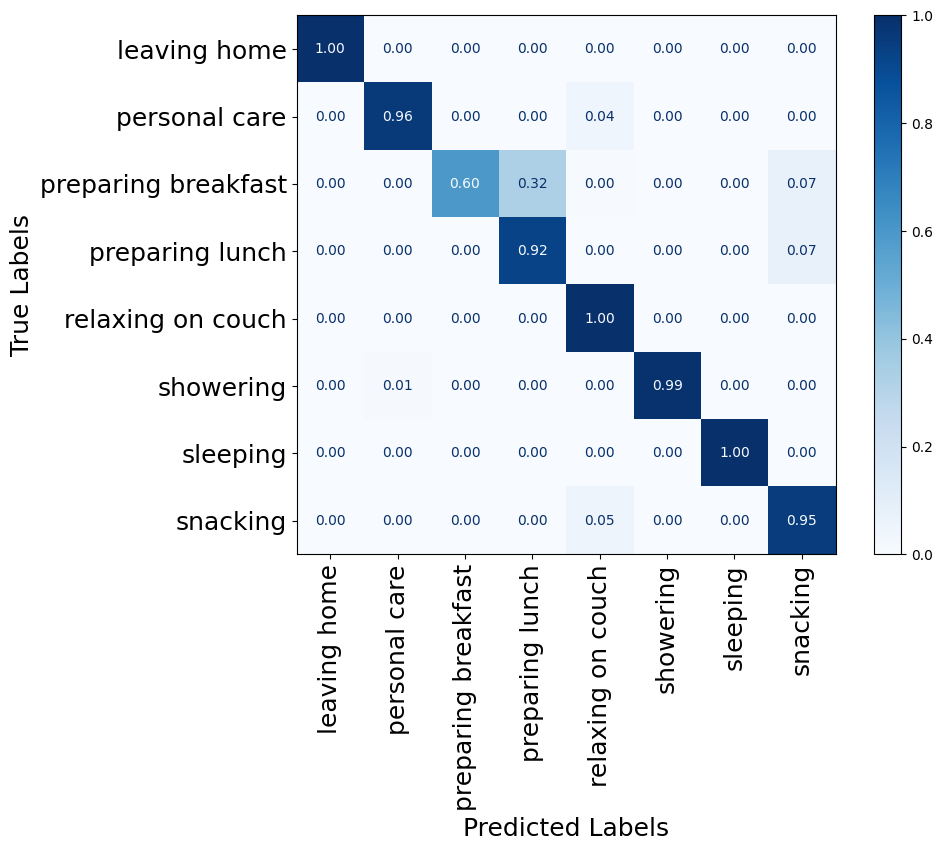}
    \caption{Home A}
  \end{subfigure}
  \hfill
  \begin{subfigure}[b]{0.49\linewidth}
    \centering
    \includegraphics[width=\linewidth]{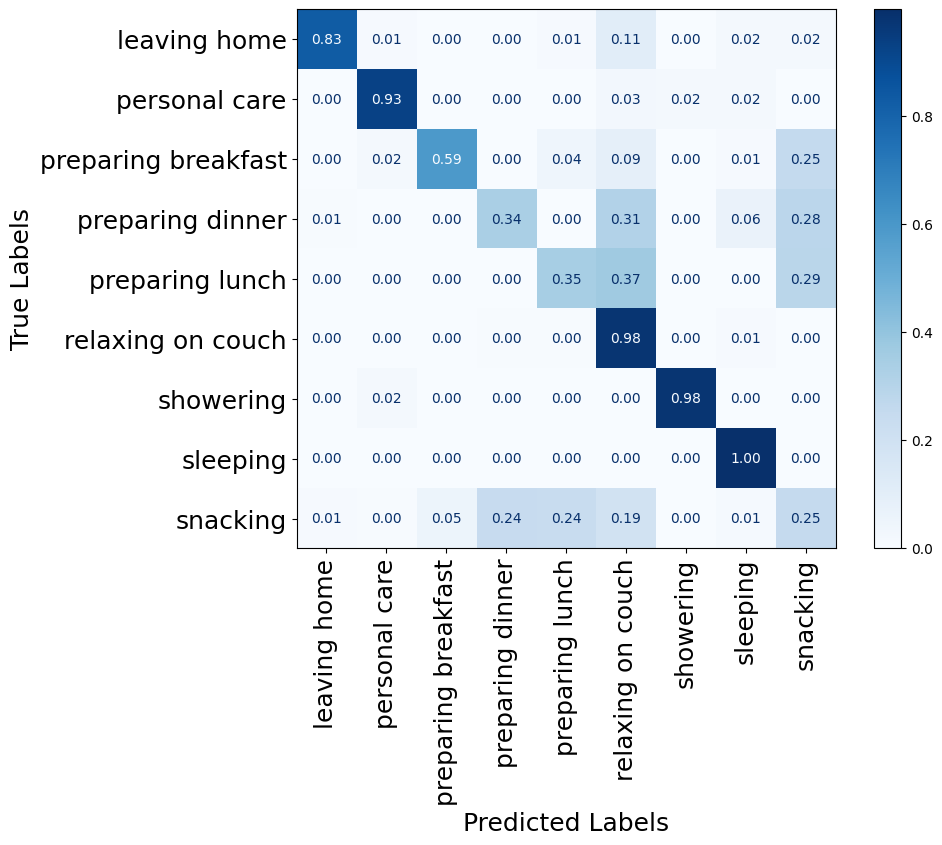}
    \caption{Home B}
  \end{subfigure}
  \caption{Zero-Shot \acronym{}: Confusion matrices on UCI ADL dataset}
  \label{fig:confusion_matrix_zero_shot_uci_adl}
\end{figure}

\begin{figure}[h!]
    \centering
    \includegraphics[width=0.55\textwidth]{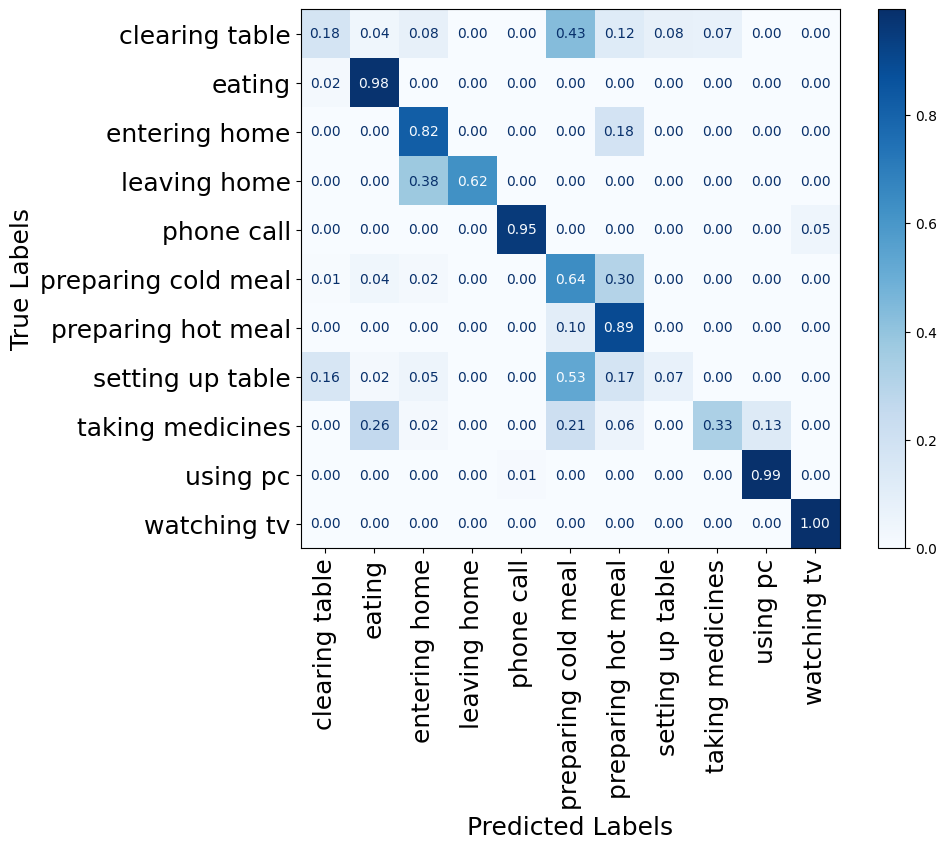}
    \caption{Zero-Shot \acronym{}: Confusion matrix on MARBLE dataset}
    \label{fig:confusion_matrix_zero_shot_marble}
\end{figure}

Figures~\ref{fig:confusion_matrix_zero_shot_uci_adl} and~\ref{fig:confusion_matrix_zero_shot_marble} depict the confusion matrices of zero-shot \acronym{}, that can be used to inspect the results more closely.
Considering the UCI-ADL dataset, the most challenging activity in both homes is \textit{snacking}. 
This phenomenon is evident in the confusion matrices depicted in Figure~\ref{fig:confusion_matrix_zero_shot_uci_adl}. Indeed, \textit{snacking} is often confused with other food-related activities like \textit{preparing breakfast} since they share similar sensor patterns.
In the Home B of the UCI ADL dataset, the \textit{relaxing on couch} activity is associated with a high number of false positives. Indeed, Figure~\ref{fig:confusion_matrix_zero_shot_uci_adl} shows that this activity is often misclassified by \acronym{} with eating-related activities and leaving home.
We believe this may happen because the subject was often eating in the living room, which is also close to the main entrance door. Hence, while supervised learning baselines could recognize these activities since they have been trained with the subject's data, the general knowledge of \acronym{} often inferred \textit{relaxing on couch} when the subject spent some time in the living room.

Considering the MARBLE dataset, we observe that \acronym{} struggles in distinguishing \textit{clearing table}, \textit{setting up the table}, and \textit{preparing a cold meal}. This is due to the lack of specific sensors characterizing these activities that involve similar actions (e.g., using the pantry, cutlery drawer, fridge). Considering \textit{taking medicines}, it is correctly classified only when the windows are associated with a state related to the interaction with the medicine cabinet. However, these interactions are usually short and might not capture the entire activity. Indeed, the subject may move to another location to take medicines, such as a dining room or kitchen, where other activities are usually performed. Finally, the misclassifications of \textit{entering home} and \textit{leaving home} are due to an insufficient sensing infrastructure to recognize these activities (i.e., no magnetic sensor on the main door). For instance, when a subject leaves home starting from the kitchen, \acronym{} sometimes mislabels it as meal preparation, while windows only including movements in the hall lead to label \textit{leaving home} as \textit{entering home} due to LLM's assumptions about the home entry points.

\begin{figure}[h!]
    \centering
    \includegraphics[width=0.35\textwidth]{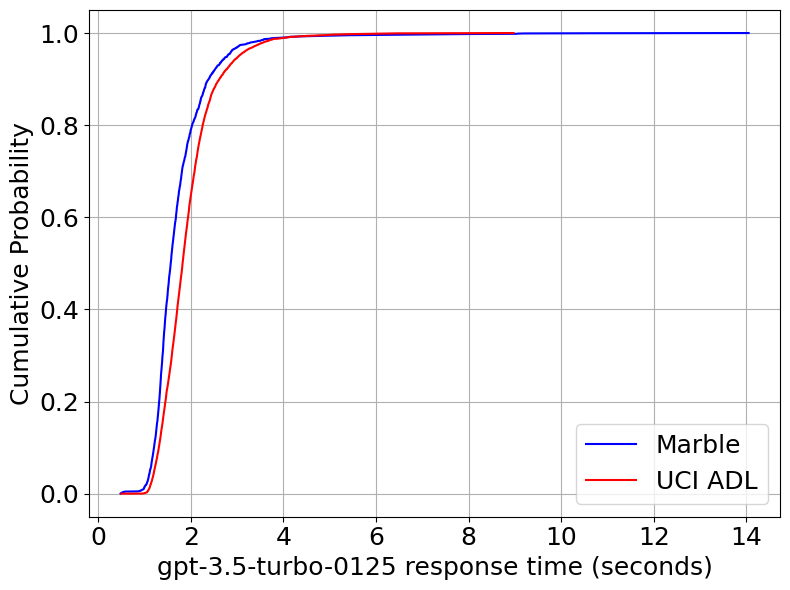}
    \caption{Zero-Shot \acronym{}: Empirical Cumulative Density Function (ECDF) of response times.}
    \label{fig:gpt_response_time_cdf}
\end{figure}

Finally, Figure~\ref{fig:gpt_response_time_cdf} shows the Empirical Cumulative Density Function (ECDF) of the response times provided by gpt-3.5 on both datasets.

We observe that $\approx 80\%$ of the responses are generated within $2$ seconds, and almost all responses for both datasets are generated within four seconds.
Only in a couple of instances, the response time was longer than $10$ seconds, probably due to networking issues.
UCI-ADL is associated with slightly higher response times due to the fact that it involves $60$ seconds windows, while  MARBLE considered $16$ seconds windows.
It is important to note that all the supervised baseline models exhibit significantly faster inference times,  processing each window on average in less than one millisecond. Among them, MLP is the fastest, with an average inference time of $0.000078$ ($\pm 0.000069$) seconds. DeepConvLSTM follows with $0.000177$ ($\pm 0.000108$) seconds, while BiLSTM is the slowest baseline with an average response time of $0.000356$ ($\pm 0.000307$) seconds. This result is expected, as the baselines are simpler models running locally, which allows them to avoid the computational overhead of large-scale LLMs and the network latency introduced by API calls.
Nevertheless, our results suggest that~\acronym{} could be also adopted for real-time ADL recognition with response times that, while not as fast as the ones reached by the baselines, are still reasonable for real-time applications.

\subsubsection{Few-shot \acronym{}} \mbox{}\\

\begin{figure}[h!]
  \centering
  \begin{subfigure}[b]{0.44\linewidth}
    \centering
    \includegraphics[width=\linewidth]{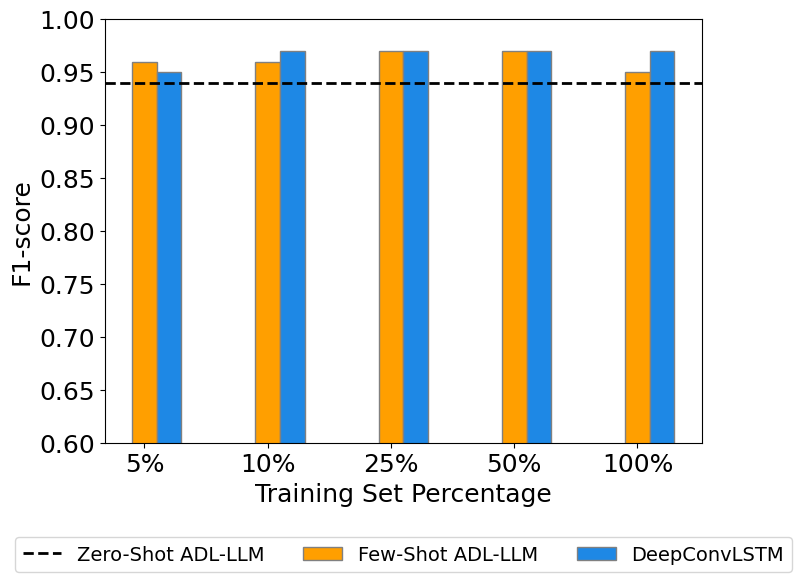}
    \caption{Home A}
    \label{subfig:barplot_home_a}
  \end{subfigure}
  \hfill
  \begin{subfigure}[b]{0.44\linewidth}
    \centering
    \includegraphics[width=\linewidth]{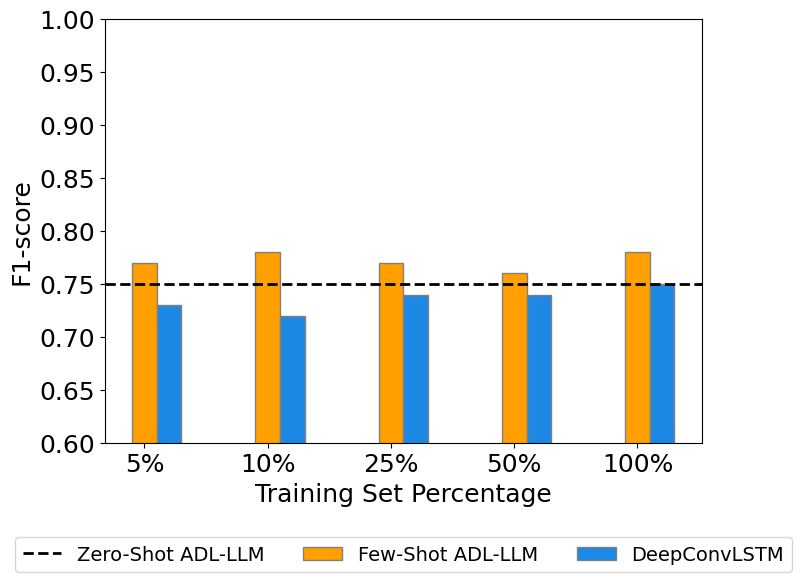}
    \caption{Home B}
    \label{subfig:barplot_home_b}
  \end{subfigure}
    \caption{UCI-ADL: Few-shot \acronym{} ($k=7$) vs DeepConvLSTM at different data scarcity scenarios.}
      \label{fig:bar_plot_uci_adl}
\end{figure}

\begin{figure}[h!]
    \centering
    \includegraphics[width=0.46\textwidth]{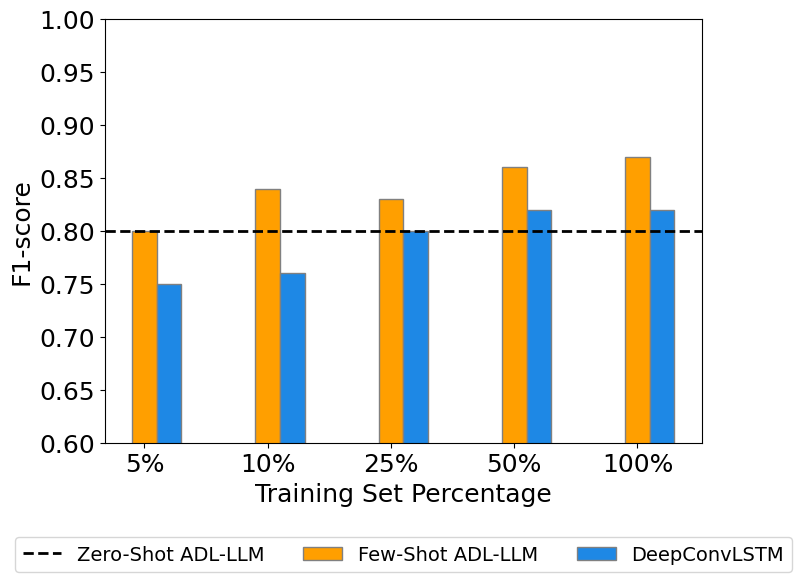}
    \caption{MARBLE: Few-shot \acronym{} ($k=7$) vs DeepConvLSTM at different data scarcity scenarios.}
    \label{fig:bar_plot_marble}
\end{figure}

In the following, we show the results of few-shot \acronym{}.
We performed the experiments by testing different values of $k$ (i.e., the number of examples), specifically $3$, $5$, $7$. Figures~\ref{fig:bar_plot_uci_adl} and~\ref{fig:bar_plot_marble} show the overall recognition rate of few-shot prompting compared to DeepConvLSTM (i.e., the best performing supervised baseline) at different data scarcity scenarios. For both datasets, we empirically determined that the best value for $k$ is $7$.

Our results show that, even when only the $5\%$ of training data is used to populate the pool of examples, few-shot \acronym{} outperforms zero-shot \acronym{}, thus showing the effectiveness of adding examples in the prompt. Except for Home A of the UCI ADL dataset, few-shot \acronym{} outperforms the supervised baseline, whose recognition rates significantly decrease as data scarcity increase.
As we mentioned before, Home A (Figure~\ref{subfig:barplot_home_a}) represents a too simple benchmark and even a small dataset leads to excellent recognition rates, and \acronym{} slightly outperforms the baseline only when considering the $5\%$ of the training set. These results confirm that, even when some labeled data are available, \acronym{} is a better choice than standard data-driven approaches.
Tables~\ref{tab:fsp_5_percent_uci_adl} and~\ref{tab:fsp_5_percent_marble} show the results at the activity level when only the $5\%$ of the training set is available. It is possible to see that increasing $k$ leads to a monotonic increase of the recognition rate.
The detailed results for the other data availability scenarios can be found in the appendix (supplemental material). 

\begin{table}[h!]
\centering
\tiny
\caption{UCI ADL: Few-Shot \acronym{} vs the supervised baselines (5\% of the training set)}
\label{tab:fsp_5_percent_uci_adl}
\begin{tabular}{|l|l|l|l|l|l|l|l|l|l|l|l|l|}
\hline
& \multicolumn{6}{c|}{ Home A } & \multicolumn{6}{c|}{ Home B} \\ \hline
& \multicolumn{3}{l|}{Few-Shot \acronym{}}& \multicolumn{3}{l|}{Supervised baselines}& \multicolumn{3}{l|}{Few-Shot \acronym{}}& \multicolumn{3}{l|}{Supervised baselines}  \\ \hline
Activity  & K=3  &  K=5  & K=7  & MLP & \multicolumn{1}{l|}{\begin{tabular}[c]{@{}l@{}} BiLSTM\end{tabular}} & \multicolumn{1}{l|}{\begin{tabular}[c]{@{}l@{}}DeepConv\\ LSTM\end{tabular}} & K=3  & K=5  & K=7  & MLP & \multicolumn{1}{l|}{\begin{tabular}[c]{@{}l@{}} BiLSTM\end{tabular}} & \multicolumn{1}{l|}{\begin{tabular}[c]{@{}l@{}}Deep\\ Conv\\ LSTM\end{tabular}} \\ \hline
Leaving home        & 1.00 & 1.00  & 1.00 & 0.47   & 0.02          & 0.97         & 0.85 & 0.85 & 0.85 & 0.30   & 0.01          & 0.85         \\ \hline
Personal care       & 0.99 & 0.99  & 0.99 & 0.90   & 0.04          & 0.94         & 0.96 & 0.96 & 0.96 & 0.92   & 0.88          & 0.95         \\ \hline
Preparing breakfast & 0.68 & 0.85  & 0.89 & 0.85   & 0.00          & 0.85         & 0.76 & 0.75 & 0.78 & 0.23   & 0.27          & 0.30         \\ \hline
Preparing dinner    & -    & -     & -    & -           & -             & -            & 0.30 & 0.34 & 0.35 & 0.13   & 0.05          & 0.32         \\ \hline
Preparing lunch     & 0.90 & 0.95  & 0.92 & 0.94   & 0.39          & 0.95         & 0.31 & 0.41 & 0.45 & 0.20   & 0.22          & 0.31         \\ \hline
Relaxing on couch   & 0.97 & 0.97  & 0.97 & 0.99   & 0.00          & 0.99         & 0.63 & 0.66 & 0.69 & 0.83   & 0.84          & 0.86         \\ \hline
Showering           & 1.00 & 1.00  & 1.00 & 0.98   & 0.01          & 0.99         & 0.95 & 0.95 & 0.95 & 0.95   & 0.87          & 0.95         \\ \hline
Sleeping            & 0.98 & 0.98  & 0.98 & 1.00   & 0.00          & 0.99         & 0.87 & 0.90 & 0.93 & 0.99   & 0.82          & 0.99         \\ \hline
Snacking            & 0.33 & 0.43  & 0.34 & 0.00   & 0.00          & 1.00         & 0.37 & 0.34 & 0.32 & 0.33   & 0.20          & 0.34         \\ \hline
\hline
\textbf{Weighted F1-score}   & 0.93 & 0.96  & 0.96 & 0.92   & 0.13          & 0.95         & 0.74 & 0.75 & 0.77 & 0.66   & 0.59          & 0.73         \\ \hline
\end{tabular}
\end{table}

\begin{table}[h!]
\tiny
\caption{MARBLE: Few-Shot \acronym{} vs the supervised baselines (5\% of the training set)}
\label{tab:fsp_5_percent_marble}
\centering
\begin{tabular}{|l|l|l|l|l|l|l|}
\hline
                    & \multicolumn{3}{l|}{Few-Shot \acronym{}}& \multicolumn{3}{l|}{Supervised baselines} \\ \hline
Activity            & K=3  & K=5  & K=7  & BiLSTM & \multicolumn{1}{l|}{\begin{tabular}[c]{@{}l@{}}Casc\\ Ens\\ BiLSTM\end{tabular}} & \multicolumn{1}{l|}{\begin{tabular}[c]{@{}l@{}}Deep\\ Conv\\ LSTM\end{tabular}} \\ \hline
Clearing table      & 0.20 & 0.21 & 0.26 & 0.00   & 0.00        & 0.01         \\ \hline
Eating              & 0.81 & 0.83 & 0.83 & 0.88   & 0.79        & 0.87         \\ \hline
Entering home       & 0.48 & 0.34 & 0.38 & 0.00   & 0.00        & 0.00         \\ \hline
Leaving home        & 0.75 & 0.48 & 0.44 & 0.00   & 0.00        & 0.00         \\ \hline
Phone call          & 0.99 & 0.99 & 0.99 & 0.54   & 0.49        & 0.78         \\ \hline
Preparing cold meal & 0.62 & 0.53 & 0.52 & 0.40   & 0.04        & 0.57         \\ \hline
Preparing hot meal  & 0.75 & 0.78 & 0.82 & 0.65   & 0.36        & 0.84         \\ \hline
Setting up table    & 0.15 & 0.32 & 0.27 & 0.14   & 0.04        & 0.46         \\ \hline
Taking medicines    & 0.22 & 0.23 & 0.30 & 0.00   & 0.00        & 0.37         \\ \hline
Using pc            & 0.96 & 0.97 & 0.97 & 0.81   & 0.62        & 0.90         \\ \hline
Watching tv         & 0.98 & 0.97 & 0.98 & 0.94   & 0.62        & 0.96         \\ \hline
\hline
\textbf{Weighted F1-score}   & 0.78 & 0.79 & 0.80 & 0.60   & 0.43        & 0.75         \\ \hline
\end{tabular}
\end{table}

\subsubsection{Alternative prompts} 
\label{subsec:alternativeprompts}
\mbox{}\\

To evaluate the robustness of our proposed prompt,
we conducted a series of experiments to assess the impact of alternative prompts to the recognition rate of zero-shot~\acronym{}. Specifically, we considered the following alternatives:

\begin{itemize}
    \item \textbf{Cleland}~\cite{cleland2024leveraging}: An approach closely related to this work. It consists of a very simple description of sensor events and minimal instructions to the LLM.
    \item \textbf{\acronym{} w/ absolute tss}: An alternative version of the Window2Text approach generating textual representations using absolute timestamps instead of relative temporal relationships (e.g.,  \textit{``after N minutes''}).
    \item  \textbf{\acronym{} w/o CoT}: Removed the Chain of Thought (CoT) reasoning from the system prompt (i.e., we removed the line \textit{``reason step by step''}).
    \item \textbf{ \acronym{} w/o Env}: An alternative system prompt where we omitted the description of the environment (i.e., the rooms and the monitored household items).
\end{itemize}

The results of this comparison are summarized in Table \ref{tab:ablation}, while examples of prompts for the alternative approaches can be found in the appendix (supplemental material).

\begin{table}[h!]
\tiny
\caption{Results for the four alternative prompts.}
\label{tab:ablation}
\centering
\begin{tabular}{|l|l|l|l|} \hline
 Prompt   &  UCI Home A & UCI Home B & Marble \\ \hline
 
  Cleland~\cite{cleland2024leveraging}  &  0.55       & 0.35       & 0.47   \\ \hline
 \acronym{} w/ absolute tss &  0.94       & 0.75       & 0.80   \\ \hline
 \acronym{} w/o CoT      &  0.94       & 0.74       & 0.80   \\ \hline
 \acronym{} w/o Env      &  0.94       & 0.74       & 0.78   \\ \hline
  \acronym{}   &  0.94       & 0.75       & 0.80   \\ \hline

\end{tabular}
\end{table}

Consistently with the results in the corresponding paper, the \textbf{Cleland}~\cite{cleland2024leveraging} approach exhibits poor results with a zero-shot strategy.  This is likely due to the simplistic nature of this approach, which yielded better results only when coupled with LLM fine-tuning.

Using absolute timestamps (\textbf{w/ absolute tss}) instead of the relative temporal relationships adopted by our Window2Text has no impact on the recognition rate, indicating that our prompting strategy provides sufficient temporal context information.

\begin{figure}[h!]
    \centering
    \includegraphics[width=0.44\textwidth]{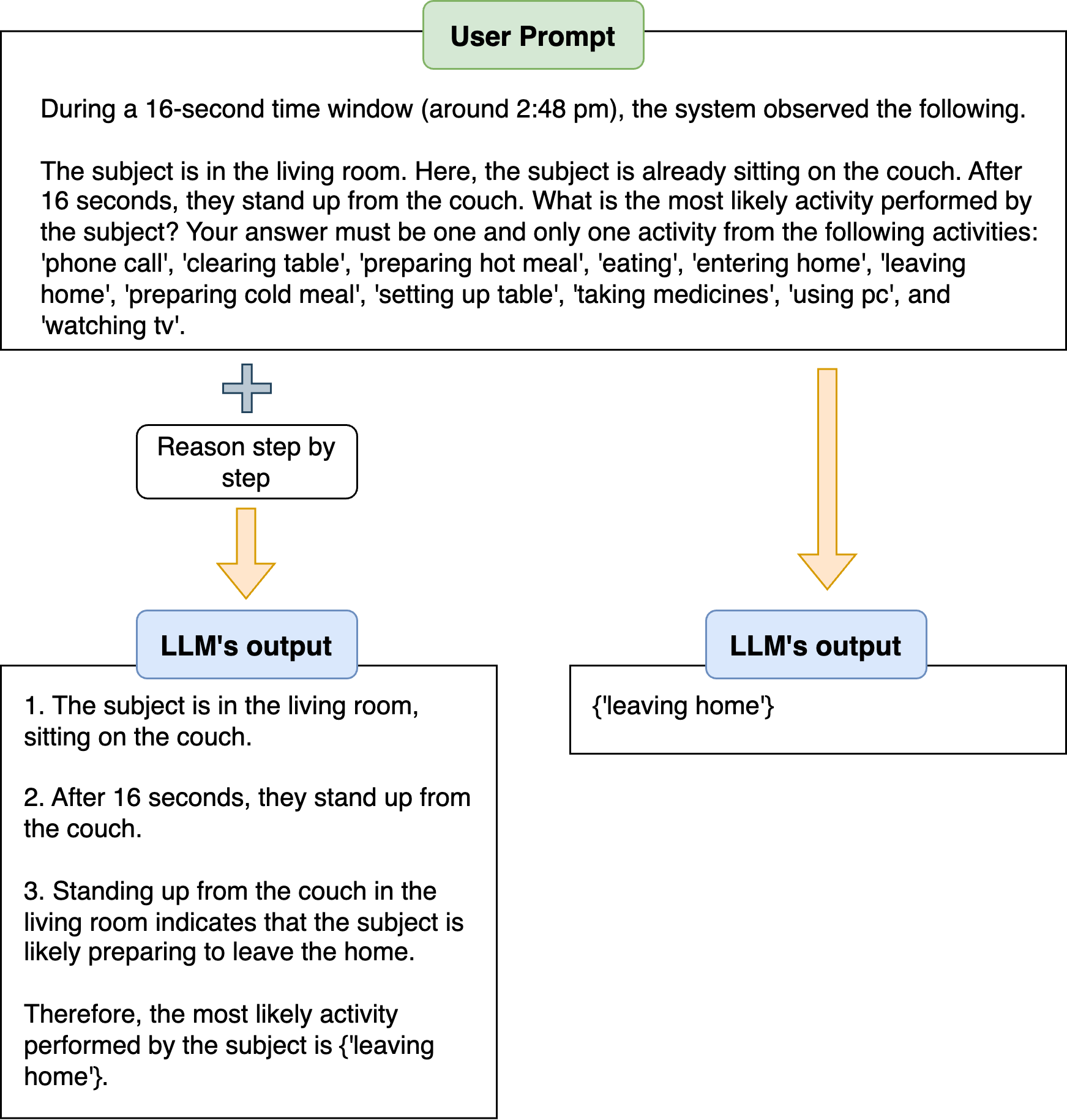}
    \caption{An example of a miss-prediction. On the left we show the output using Chain-of-Thought (CoT), while on the right the output without CoT. In the first case, the output is more interpretable, revealing that the model erroneously infers the subject is leaving the home simply because they stood up from the couch.}
    \label{fig:ablation_no_cot_example}
\end{figure}

An unexpected result is that removing CoT (\textbf{w/o CoT}) has almost no impact on the recognition rate, while its adoption is known for improving the LLM's output in many other domains~\cite{wei2022chain}.
This may be due to the nature of the considered domain, but larger-scale experiments will be needed to confirm this insight.
However, an advantage of keeping CoT in our prompt is that it improves the output interpretability. Figure \ref{fig:ablation_no_cot_example} illustrates an instance of a window that is incorrectly classified (with the same activity) both with and without CoT. In the first case, the model offers an insight into the LLM’s “reasoning process” that leads to the incorrect output, while in the second, it only provides the predicted label without any explanation, making it more challenging to diagnose the underlying causes of the model’s mistakes.

Finally, excluding the smart home description from the system prompt (\textbf{w/o env}) leads to a slight drop in recognition accuracy. Indeed, in a few instances, the model leverages contextual knowledge about the monitored household items. 
For instance, consider the following example: 

\begin{center}
\textit{During a 16-second time window (around 5:11 PM), the system observed the following. The subject is in the kitchen. Here, the pantry drawer is already open. After 2 seconds, they close the pantry drawer.}
\end{center}

We observe that in this case, \acronym{} takes into consideration the fact that the stove is one of the monitored appliances (even if not used by the subject in the current window) to correctly predict the "preparing a cold meal" activity. Indeed, the LLM motivated that the current activity can not be \textit{preparing a hot meal} with the following argument:

\begin{center}
    \textit{``[...] Preparing hot meal: This could involve using ingredients from the pantry, but we don't have evidence of cooking or using the stove. [...]''}
\end{center}\

On the other hand, if the environment's description is omitted, the response to the same input becomes:

\textit{[...] Since the pantry is typically used for storing food items, and the subject's action of closing the drawer implies they were likely preparing something, the most fitting activity is 'preparing hot meal' or 'preparing cold meal'. However, without additional context indicating the type of meal, 'preparing hot meal' is a common activity associated with kitchen use. Thus, the most likely activity performed by the subject is: \{'preparing hot meal'\}. [...]}

\section{Discussion}
\label{sec:discussion}

\subsection{Segmentation}

In this work, we segmented sensor data using fixed-size windows, which is a standard approach in the literature. However, while this approach makes sense to train machine learning classifiers and find low-level patterns, this may be less suitable for LLMs. Indeed, fixed-size segments capture only a small portion of the activity, and they often include transitions between different activities.

We will explore dynamic segmentation strategies based on change-point detection~\cite{aminikhanghahi2017survey}. Indeed, dynamic segmentation can be leveraged to detect significant changes in sensor data distribution that are likely associated with changes in the performed activities~\cite{aminikhanghahi2017using}. We believe that dynamic segmentation would allow LLMs to process more meaningful information to reliably perform ADL recognition.

\subsection{Open-world HAR}
\label{sec:openworld}

For the sake of this work, we constrained the LLM in predicting the most likely activity among a set of candidates. This is the most common approach in the ADLs recognition literature.
However, this is a limitation, since in real-life deployments we can not foresee in advance all the possible activities carried out by the subject.
Moreover, the answers from the model may ``overfit'' on the specific activity sets, especially when there is a one-to-one mapping between activities in the set and sensor events.

We will explore whether the flexibility of pre-trained LLMs can also be effective in open-world HAR, where there is no fixed set of activities to consider. One of the main challenges will be that the LLM may provide different labels for the same activity (e.g., eating may also be recognized as dining, snacking, having lunch, etc.). Hence, we believe that in such settings it will be necessary to semantically cluster the predicted labels.

\subsection{Cloud vs Local LLM processing}

In our experiments, we adopted cloud-based LLM models due to their effectiveness and to simplify the implementation. However, outsourcing information about personal activities to such untrusted third parties raises several privacy concerns. Indeed, personal data about human activities may reveal private habits and health issues~\cite{bettini2015privacy}.
Moreover, since we use the LLM for each sensor data time window, \acronym{} may be too costly when using cloud-based LLMs.

Our vision is that open-source LLMs should be used in real-world implementations of \acronym{}. Since the deployment of such models may be prohibitive in smart-home gateways, we believe that they should run on high-performance computing machines in a trusted domain (e.g., in a telemedicine infrastructure considering a medical application). 
When this is not possible, compressed versions (e.g., quantized versions and/or with reduced number of parameters) may be even deployed in the home domain (e.g., in a home gateway powered with a GPU). 
However, since these models are currently less effective, we expect a significant drop in the recognition rate.

We evaluated the impact of smaller open source LLMs on \acronym{} by performing preliminary experiments by deploying the Llama-2 model on our infrastructure. Specifically, we used the \textit{llama-2-13b-chat-hf} model through HuggingFace~\footnote{\url{https://huggingface.co/meta-llama/Llama-2-13b-chat-hf}}.  
We ran these experiments on a Ubuntu server with a partitioned NVIDIA A100 GPU, with $42$GB of RAM, and an AMD EPYC 8-core CPU. Table~\ref{tab:llama2_marble} compares zero-shot \acronym{} using Llama-2 and gpt3.5 on the MARBLE dataset.

\begin{table}[h!]
\tiny
\caption{Zero-shot \acronym{} (MARBLE): \textit{llama-2-13b-chat-hf} vs. \textit{gpt-3.5-turbo-0125}}
\label{tab:llama2_marble}
\begin{tabular}{|l|c|c|}
\hline
Activity            & llama-2-13b-chat-hf & gpt-3.5-turbo-0125 \\ \hline
Clearing table      & 0.23                & 0.25               \\ \hline
Eating              & 0.60                & 0.91              \\ \hline
Entering home       & 0.58                & 0.41               \\ \hline
Leaving home        & 0.86                & 0.71               \\ \hline
Phone call          & 0.99                & 0.97               \\ \hline
Preparing cold meal & 0.00                & 0.51               \\ \hline
Preparing hot meal  & 0.61                & 0.82               \\ \hline
Setting up table    & 0.00                & 0.13               \\ \hline
Taking medicines    & 0.34                & 0.47               \\ \hline
Using pc            & 0.98                & 0.97               \\ \hline
Watching tv         & 0.96                & 0.97               \\ \hline
\hline
\textbf{Weighted F1-score}   &  0.68                   & 0.80               \\ \hline
\end{tabular}
\end{table}

As expected, \textit{gpt-3.5-turbo-0125} shows superior performance than \textit{llama-2-13b-chat-hf}. 
Specifically, Llama-2 struggles even more in recognizing the activities lacking distinct sensor inputs like preparing a cold meal, setting up the table, and clearing the table.
Additionally, \textit{llama-2-13b-chat-hf} was not accurate in providing the output activity in the required format. As we described in Section~\ref{subsec:sysprompt}, the system prompt instructs the system to enclose the most likely activity within curly braces. While \textit{gpt-3.5-turbo-0125} always complied, \textit{llama-2-13b-chat-hf} often did not adhere to this instruction. This required us to find ad-hoc solutions to extract activity labels when the output format was different.
Furthermore, \textit{llama-2-13b-chat-hf} tended to output activity labels not even semantically close to the predefined activity sets, such as \textit{`moving around the house'}.
Hence, in the future, we will investigate alternative prompting and personalization strategies (e.g., fine-tuning) for these smaller models.

\subsection{Hallucinations}

LLMs are subject to hallucinations, that is the generation of content that may be irrelevant, made-up, or inconsistent with the input data. 
In our use of LLMs, a hallucination may lead to a wrong prediction of the activity performed by the subject. It may also lead to an output that is not compliant with the instructions (e.g., predicting an activity not included in the given set or using a different format).
The impact of the first is similar to the one due to miss-predictions obtained with classic data-driven models. In our results, the F1 score captures this impact together with other possible factors like for example the effectiveness of the pre-processing.

Hallucinations leading to non-compliant outputs never occurred with the advanced models we adopted in our experiments, while they may occur when using simpler and smaller LLMs (as we previously discussed in Section~\ref{sec:openworld}). 
To avoid this problem, a test can be added to check whether the output activity is in the required set and, if necessary, ask the LLM to correct its prediction.

In general, hallucination mitigation actions can be taken. For example,  the \acronym{}'s system prompt adopts the Chain of Thought (CoT) that is known to mitigate the problem, even though we did not observe an advantage in terms of recognition rates in our experiments besides improving interpretability. Few-shot prompting also reduces the risk of hallucination. In future work, we will explore alternative approaches (e.g., ensembles of different LLMs).

\subsection{Impact of LLMs evolution}

While we executed all of our experiments running gpt-3.5 due to its cost-effectiveness and accuracy, one may argue that more advanced LLM models may lead to even better results. In the following, we show preliminary experiments of zero-shot \acronym{} using the more advanced \textit{gpt-4o} model on the MARBLE dataset.
Table~\ref{tab:gpt_4o_marble} shows that \textit{gpt-4o} improves the F1-score by $+4\%$ over \textit{gpt-3.5-turbo-0125}. 

\begin{table}[h!]
\tiny
\caption{Zero-Shot \acronym{} (MARBLE): \textit{gpt-4o} vs. \textit{gpt-3.5-turbo-0125}}
\label{tab:gpt_4o_marble}
\centering
\begin{tabular}{|l|c|c|}
\hline
Activity            & gpt-4o & gpt-3.5-turbo-0125 \\ \hline
Clearing table      & 0.23   & 0.25               \\ \hline
Eating              & 0.91   & 0.91              \\ \hline
Entering home       & 0.60   & 0.41               \\ \hline
Leaving home        & 0.74   & 0.71               \\ \hline
Phone call          & 0.99   & 0.97               \\ \hline
Preparing cold meal & 0.55   & 0.51               \\ \hline
Preparing hot meal  & 0.85   & 0.82               \\ \hline
Setting up table    & 0.49   & 0.13               \\ \hline
Taking medicines    & 0.56   & 0.47               \\ \hline
Using pc            & 0.99   & 0.97               \\ \hline
Watching tv         & 0.99   & 0.97               \\ \hline
\hline
\textbf{Weighted F1-score}   & 0.84   & 0.80               \\ \hline
\end{tabular}
\end{table}

These results indicate that using more advanced LLMs, it is possible to even surpass DeepConvLSTM trained on the whole training set (see Table~\ref{tab:zs-marble}). However, this performance gain comes with a significant increase in costs, since at the time of writing running experiments with \textit{gpt-4o} is $10$ times more expensive than with \textit{gpt-3.5-turbo-0125}. Moreover, this advanced model was significantly slower in generating the output. This experiment confirms that LLMs are zero-shot ADL recognition models, but doing it accurately by minimizing costs and ensuring privacy is still an open challenge.

\subsection{Robustness to noisy sensor data}

Similarly to related work in this area, we test our algorithms on public datasets that usually have undergone a data cleaning pre-processing.
Sensors in real-world deployments may generate incorrect/noisy data or experience faults leading to missing data in the resulting data stream. This problem negatively impacts any HAR system independently of the underlying model: from classic data-driven solutions to our \acronym{}. 
In data-driven supervised approaches, the larger the available training set the more resilience to data noise is achieved. 
However, real-world scenarios are often characterized by limited (or none) labeled data availability.
In the literature, common knowledge is known for improving the robustness to noise of HAR models in labeled data scarcity settings~\cite{arrotta2024semantic}. Considering \acronym{}, the common knowledge encoded in the LLM should help to identify activities even when some of the sensors are noisy or fail to provide data.
%
Nonetheless, when data is too noisy or the amount of missing information is significant, both \acronym{} and classic data-driven solutions would lead to poor results.
%
In future work, we may perform experiments to evaluate the actual robustness of LLMs against sensor noise, by leveraging the real-world smart home data that we are currently collecting from the homes of elderly subjects in the context of a project in collaboration with neurologists. We may also design possible prompting strategies to make the LLM aware that the processed windows may be noisy.



\section{Conclusion and Future Work}
In this paper, we presented \acronym{}: a zero-shot method for ADL recognition in smart-home environments. We showed that LLMs can reach recognition rates close to the ones of supervised approaches, with the benefit of not requiring labeled data.
While we evaluated \acronym{} on public datasets, we intend to investigate its effectiveness in the specific medical context of neurology, considering data from real subjects at risk of cognitive decline.
We are currently collaborating with clinicians in the SERENADE project\footnote{\url{https://ecare.unimi.it/pilots/serenade/}}, collecting data from the smart homes of a significantly large group of subjects, mostly living alone and diagnosed with Mild Cognitive Impairment. Our long-term objective in this project is a fine-grained recognition of the onset of cognitive decline. The sensor data that we are collecting in these homes is unlabeled, and we plan to adopt zero-shot \acronym{} to capture the performed ADLs in this setting and work from there on detecting behavioral changes and anomalies under the guidance of clinicians. 



\begin{acks}
This work was supported in part by projects MUSA, FAIR, and SERICS under the NRRP MUR program funded by the EU-NGEU. Views and opinions expressed are however those of the authors only and do not necessarily reflect those of the European Union or the Italian MUR. Neither the European Union nor the Italian MUR can be held responsible for them.
\end{acks}

\bibliographystyle{ACM-Reference-Format}
\bibliography{references}

\newpage

\end{document}


\title{Appendix: Large Language Models are Zero-Shot Recognizers for Activities of Daily Living}

\author{Gabriele Civitarese}
\email{gabriele.civitarese@unimi.it}
\author{Michele Fiori}
\email{michele.fiori@unimi.it}
\author{Priyankar Choudhary}
\email{priyankar.choudhary@unimi.it}
\author{Claudio Bettini}
\email{claudio.bettini@unimi.it}
\affiliation{%
  \institution{EveryWare Lab, Dept. of Computer Science, University of Milan}
  \country{Italy}
}

\maketitle

\appendix

\section{Detailed results at different data availability scenarios}
\label{sec:app}

In the following, we report the results of few-shot prompting (and the baselines) at different data availability scenarios: $100\%$, $50\%$, $25\%$, $10\%$ of the training set.

\begin{table}[h]
\tiny
\caption{UCI ADL: Few-Shot \acronym{} vs the supervised baselines ($100\%$ of the training set)}
\label{tab:fsp_100_percent_uci_adl}
\begin{tabular}{|l|l|l|l|l|l|l|l|l|l|l|l|l|}
\hline
& \multicolumn{6}{c|}{ Home A } & \multicolumn{6}{c|}{ Home B} \\ \hline
& \multicolumn{3}{l|}{Few-Shot \acronym{}}& \multicolumn{3}{l|}{Supervised baselines}& \multicolumn{3}{l|}{Few-Shot-Prompting}& \multicolumn{3}{l|}{Supervised baselines}  \\ \hline
Activity            & K=3   & K=5  & K=7  & {MLP} & \multicolumn{1}{l|}{\begin{tabular}[c]{@{}l@{}} BiLSTM\end{tabular}} & \multicolumn{1}{l|}{\begin{tabular}[c]{@{}l@{}}DeepConv\\ LSTM\end{tabular}} & K=3   & K=5  & K=7  & {MLP} & \multicolumn{1}{l|}{\begin{tabular}[c]{@{}l@{}}BiLSTM\end{tabular}} & \multicolumn{1}{l|}{\begin{tabular}[c]{@{}l@{}}DeepConv\\ LSTM\end{tabular}} \\ \hline
Leaving home        & 1.00  & 1.00 & 1.00 & {0.97}   & 0.95          & 0.99         & 0.90  & 0.90 & 0.90 & {0.85}   & 0.84          & 0.88         \\ \hline
Personal care       & 1.00  & 1.00 & 0.99 & {0.99}   & 0.98          & 0.99         & 0.95  & 0.96 & 0.96 & {0.94}   & 0.92          & 0.93         \\ \hline
Preparing breakfast & 0.81  & 0.89 & 0.94 & {0.92}   & 0.90          & 0.94         & 0.71  & 0.83 & 0.81 & {0.29}   & 0.35          & 0.39         \\ \hline
Preparing dinner    & -     & -    & -    & -           & -             & -            & 0.29  & 0.39 & 0.40 & {0.16}   & 0.06          & 0.13         \\ \hline
Preparing lunch     & 0.94  & 0.96 & 0.97 & {0.96}   & 0.95          & 0.96         & 0.38  & 0.45 & 0.50 & {0.30}   & 0.35          & 0.41         \\ \hline
Relaxing on couch   & 0.86  & 0.87 & 0.87 & {1.00}   & 1.00          & 1.00         & 0.61  & 0.64 & 0.71 & {0.86}   & 0.85          & 0.87         \\ \hline
Showering           & 1.00  & 1.00 & 1.00 & {0.99}   & 0.99          & 1.00         & 0.92  & 0.94 & 0.95 & {0.95}   & 0.93          & 0.96         \\ \hline
Sleeping            & 0.89  & 0.90 & 0.90 & {1.00}   & 0.99          & 1.00         & 0.84  & 0.82 & 0.89 & {0.99}   & 0.99          & 0.99         \\ \hline
Snacking            & 0.41  & 0.54 & 0.60 & {0.00}   & 0.04          & 0.00         & 0.35  & 0.39 & 0.37 & {0.39}   & 0.37          & 0.40         \\ \hline
Weighted F1-score   & 0.93  & 0.95 & 0.95 & {0.96}   & 0.96          & 0.97         & 0.73  & 0.76 & 0.78 & {0.73}   & 0.72          & 0.75         \\ \hline
\end{tabular}
\end{table}


\begin{table}[h]
\tiny
\caption{MARBLE: Few-Shot \acronym{} vs the supervised baselines ($100\%$ of the training set)}
\label{tab:fsp_100_percent_marble}

\begin{tabular}{|l|l|l|l|l|l|l|}
\hline
& \multicolumn{3}{l|}{Few-Shot \acronym{}}& \multicolumn{3}{l|}{Supervised baselines} \\ \hline
Activity            & K=3  & K=5  & K=7  & {MLP} & \multicolumn{1}{l|}{\begin{tabular}[c]{@{}l@{}}BiLSTM\end{tabular}} & \multicolumn{1}{l|}{\begin{tabular}[c]{@{}l@{}}DeepConv\\ LSTM\end{tabular}} \\ \hline
Clearing table      & 0.51 & 0.62 & 0.56 & {0.40}   & 0.44        & 0.55         \\ \hline
Eating              & 0.88 & 0.87 & 0.88 & {0.91}   & 0.91        & 0.91         \\ \hline
Entering home       & 0.69 & 0.65 & 0.73 & {0.00}   & 0.00        & 0.13         \\ \hline
Leaving home        & 0.77 & 0.77 & 0.77 & {0.00}   & 0.00        & 0.00         \\ \hline
Phone call          & 0.99 & 0.99 & 0.99 & {0.80}   & 0.76        & 0.81         \\ \hline
Preparing cold meal & 0.68 & 0.69 & 0.71 & {0.70}   & 0.63        & 0.73         \\ \hline
Preparing hot meal  & 0.81 & 0.86 & 0.86 & {0.89}   & 0.82        & 0.90         \\ \hline
Setting up table    & 0.44 & 0.50 & 0.52 & {0.58}   & 0.55        & 0.57         \\ \hline
Taking medicines    & 0.59 & 0.60 & 0.58 & {0.57}   & 0.54        & 0.59         \\ \hline
Using PC            & 0.98 & 0.99 & 0.99 & {0.91}   & 0.91        & 0.91         \\ \hline
Watching TV         & 0.99 & 0.99 & 0.99 & {0.95}   & 0.94        & 0.96         \\ \hline
Weighted F1-score   & 0.85 & 0.87 & 0.87 & {0.81}   & 0.78        & 0.82         \\ \hline
\end{tabular}
\end{table}



\begin{table}[h]
\tiny
\caption{UCI ADL: Few-Shot \acronym{} vs the supervised baselines ($50\%$ of the training set)}
\label{tab:fsp_50_percent_uci_adl}
\begin{tabular}{|l|l|l|l|l|l|l|l|l|l|l|l|l|}
\hline
& \multicolumn{6}{c|}{ Home A } & \multicolumn{6}{c|}{ Home B} \\ \hline
& \multicolumn{3}{l|}{Few-Shot \acronym{}}& \multicolumn{3}{l|}{Supervised baselines}& \multicolumn{3}{l|}{Few-Shot \acronym{}}& \multicolumn{3}{l|}{Supervised baselines}  \\ \hline
Activity            & K=3   & K=5  & K=7  & {MLP} & \multicolumn{1}{l|}{\begin{tabular}[c]{@{}l@{}}BiLSTM\end{tabular}} & \multicolumn{1}{l|}{\begin{tabular}[c]{@{}l@{}}DeepConv\\ LSTM\end{tabular}} & K=3   & K=5  & K=7  & {MLP} & \multicolumn{1}{l|}{\begin{tabular}[c]{@{}l@{}}BiLSTM\end{tabular}} & \multicolumn{1}{l|}{\begin{tabular}[c]{@{}l@{}}DeepConv\\ LSTM\end{tabular}} \\ \hline
Leaving home        & 1.00  & 1.00 & 1.00 & {0.94}   & 0.93          & 0.99         & 0.89  & 0.88 & 0.89 & {0.84}   & 0.73          &0.86          \\ \hline
Personal care       & 0.99  & 1.00 & 1.00 & {0.99}   & 0.96          & 0.99         & 0.93  & 0.95 & 0.96 & {0.94}   & 0.92          &0.93          \\ \hline
Preparing breakfast & 0.76  & 0.86 & 0.91 & {0.93}   & 0.88          & 0.92         & 0.80  & 0.82 & 0.83 & {0.20}   & 0.36          &0.38          \\ \hline
Preparing dinner    & -     & -    & -    & -           & -             & -            & 0.39  & 0.40 & 0.39 & {0.14}   & 0.05          &0.14          \\ \hline
Preparing lunch     & 0.91  & 0.95 & 0.97 & {0.96}   & 0.95          & 0.96         & 0.39  & 0.43 & 0.44 & {0.26}   & 0.32          &0.39          \\ \hline
Relaxing on couch   & 0.83  & 0.93 & 0.98 & {1.00}   & 1.00          & 0.99         & 0.64  & 0.64 & 0.65 & {0.86}   & 0.84          &0.87          \\ \hline
Showering           & 0.99  & 1.00 & 1.00 & {0.99}   & 0.98          & 0.99         & 0.95  & 0.95 & 0.96 & {0.92}   & 0.92          &0.95          \\ \hline
Sleeping            & 0.90  & 0.94 & 0.98 & {1.00}   & 0.99          & 1.00         & 0.81  & 0.85 & 0.85 & {0.99}   & 0.99          &0.99          \\ \hline
Snacking            & 0.36  & 0.56 & 0.69 & {0.13}   & 0.04          & 0.04         & 0.38  & 0.34 & 0.35 & {0.39}   & 0.25          &0.38          \\ \hline
Weighted F1-score   & 0.91  & 0.95 & 0.97 & {0.97}   & 0.95          & 0.97         & 0.75  & 0.76 & 0.76 & {0.71}   & 0.70          &0.74          \\ \hline
\end{tabular}
\end{table}


\begin{table}[h]
\tiny
\caption{UCI ADL: Few-Shot \acronym{} vs the supervised baselines ($50\%$ of the training set)}
\label{tab:fsp_50_percent_marble}

\begin{tabular}{|l|l|l|l|l|l|l|}
\hline & \multicolumn{3}{l|}{Few-Shot-Prompting}& \multicolumn{3}{l|}{Supervised baselines} \\ \hline
Activity            & K=3  & K=5  & K=7  & {MLP} & \multicolumn{1}{l|}{\begin{tabular}[c]{@{}l@{}}BiLSTM\end{tabular}} & \multicolumn{1}{l|}{\begin{tabular}[c]{@{}l@{}}DeepConv\\ LSTM\end{tabular}} \\ \hline
Clearing table      & 0.44 & 0.49 & 0.57 & {0.37}   & 0.29        & 0.46         \\ \hline
Eating              & 0.87 & 0.90 & 0.91 & {0.91}   & 0.91        & 0.91         \\ \hline
Entering home       & 0.60 & 0.57 & 0.57 & {0.04}   & 0.00        & 0.16         \\ \hline
Leaving home        & 0.80 & 0.75 & 0.86 & {0.29}   & 0.00        & 0.39         \\ \hline
Phone call          & 0.99 & 0.99 & 0.99 & {0.80}   & 0.76        & 0.81         \\ \hline
Preparing cold meal & 0.67 & 0.64 & 0.69 & {0.68}   & 0.61        & 0.71         \\ \hline
Preparing hot meal  & 0.84 & 0.83 & 0.85 & {0.88}   & 0.83        & 0.89         \\ \hline
Setting up table    & 0.37 & 0.52 & 0.47 & {0.57}   & 0.52        & 0.62         \\ \hline
Taking medicines    & 0.42 & 0.43 & 0.42 & {0.53}   & 0.53        & 0.58         \\ \hline
Using PC            & 0.99 & 0.99 & 0.99 & {0.91}   & 0.91        & 0.90         \\ \hline
Watching TV         & 0.99 & 0.99 & 0.99 & {0.96}   & 0.95        & 0.97         \\ \hline
Weighted F1-score   & 0.84 & 0.85 & 0.86 & {0.80}   & 0.77        & 0.82         \\ \hline
\end{tabular}
\end{table}


\begin{table}[h]
\tiny
\caption{UCI ADL: Few-Shot \acronym{} vs the supervised baselines ($25\%$ of the training set)}
\label{tab:fsp_25_percent_uci_adl}
\begin{tabular}{|l|l|l|l|l|l|l|l|l|l|l|l|l|}
\hline
& \multicolumn{6}{c|}{ Home A } & \multicolumn{6}{c|}{ Home B} \\ \hline
& \multicolumn{3}{l|}{Few-Shot \acronym{}}& \multicolumn{3}{l|}{Supervised baselines}& \multicolumn{3}{l|}{Few-Shot \acronym{}}& \multicolumn{3}{l|}{Supervised baselines}  \\ \hline
Activity            & K=3   & K=5  & K=7  & {MLP} & \multicolumn{1}{l|}{\begin{tabular}[c]{@{}l@{}}BiLSTM\end{tabular}} & \multicolumn{1}{l|}{\begin{tabular}[c]{@{}l@{}}DeepConv\\ LSTM\end{tabular}} & K=3  & K=5  & K=7  & {MLP} & \multicolumn{1}{l|}{\begin{tabular}[c]{@{}l@{}}BiLSTM\end{tabular}} & \multicolumn{1}{l|}{\begin{tabular}[c]{@{}l@{}}DeepConv\\ LSTM\end{tabular}} \\ \hline
Leaving home        & 1.00  & 1.00 & 1.00 & {0.83}   & 0.79          & 0.98         & 0.86 & 0.88 & 0.88 & {0.80}   & 0.64          & 0.84         \\ \hline
Personal care       & 0.99  & 0.99 & 1.00 & {0.98}   & 0.96          & 0.99         & 0.95 & 0.95 & 0.96 & {0.93}   & 0.90          & 0.94         \\ \hline
Preparing breakfast & 0.77  & 0.82 & 0.86 & {0.91}   & 0.85          & 0.91         & 0.73 & 0.78 & 0.78 & {0.27}   & 0.36          & 0.38         \\ \hline
Preparing dinner    & -     & -    & -    & -           & -             & -            & 0.33 & 0.37 & 0.42 & {0.18}   & 0.04          & 0.25         \\ \hline
Preparing lunch     & 0.92  & 0.95 & 0.95 & {0.94}   & 0.94          & 0.95         & 0.39 & 0.43 & 0.44 & {0.17}   & 0.16          & 0.18         \\ \hline
Relaxing on couch   & 0.85  & 0.95 & 0.98 & {1.00}   & 1.00          & 0.99         & 0.67 & 0.67 & 0.71 & {0.85}   & 0.84          & 0.86         \\ \hline
Showering           & 0.99  & 1.00 & 1.00 & {1.00}   & 0.98          & 1.00         & 0.95 & 0.94 & 0.96 & {0.93}   & 0.90          & 0.95         \\ \hline
Sleeping            & 0.89  & 0.96 & 0.99 & {1.00}   & 0.99          & 1.00         & 0.88 & 0.86 & 0.88 & {0.99}   & 0.98          & 0.99         \\ \hline
Snacking            & 0.37  & 0.58 & 0.61 & {0.00}   & 0.02          & 0.00         & 0.36 & 0.35 & 0.34 & {0.40}   & 0.29          & 0.44         \\ \hline
Weighted F1-score   & 0.91  & 0.95 & 0.97 & {0.96}   & 0.94          & 0.97         & 0.75 & 0.76 & 0.77 & {0.71}   & 0.67          & 0.74         \\ \hline
\end{tabular}
\end{table}


\begin{table}[h]
\tiny
\caption{MARBLE: Few-Shot \acronym{} vs the supervised baselines ($25\%$ of the training set)}
\label{tab:fsp_25_percent_marble}

\begin{tabular}{|l|l|l|l|l|l|l|}
\hline
                    & \multicolumn{3}{l|}{Few-Shot \acronym{}}& \multicolumn{3}{l|}{Supervised baselines} \\ \hline
Activity            & K=3  & K=5  & K=7  & {MLP} & \multicolumn{1}{l|}{\begin{tabular}[c]{@{}l@{}}BiLSTM\end{tabular}} & \multicolumn{1}{l|}{\begin{tabular}[c]{@{}l@{}}DeepConv\\ LSTM\end{tabular}} \\ \hline
Clearing table      & 0.39 & 0.44 & 0.41 & {0.39}   & 0.30        & 0.49         \\ \hline
Eating              & 0.89 & 0.89 & 0.89 & {0.90}   & 0.91        & 0.90         \\ \hline
Entering home       & 0.32 & 0.59 & 0.48 & {0.00}   & 0.00        & 0.00         \\ \hline
Leaving home        & 0.88 & 0.88 & 0.93 & {0.11}   & 0.00        & 0.14         \\ \hline
Phone call          & 0.98 & 0.99 & 0.99 & {0.76}   & 0.71        & 0.78         \\ \hline
Preparing cold meal & 0.64 & 0.60 & 0.57 & {0.68}   & 0.61        & 0.69         \\ \hline
Preparing hot meal  & 0.78 & 0.81 & 0.80 & {0.86}   & 0.79        & 0.86         \\ \hline
Setting up table    & 0.46 & 0.47 & 0.40 & {0.54}   & 0.36        & 0.58         \\ \hline
Taking medicines    & 0.49 & 0.50 & 0.47 & {0.44}   & 0.45        & 0.54         \\ \hline
Using pc            & 0.98 & 0.98 & 0.99 & {0.90}   & 0.91        & 0.91         \\ \hline
Watching tv         & 0.99 & 0.99 & 0.98 & {0.94}   & 0.94        & 0.95         \\ \hline
Weighted F1-score   & 0.83 & 0.84 & 0.83 & {0.78}   & 0.72        & 0.80         \\ \hline
\end{tabular}
\end{table}


\begin{table}[]
\tiny
\caption{UCI ADL: Few-Shot \acronym{} vs the supervised baselines ($10\%$ of the training set)}
\label{tab:fsp_10_percent_uci_adl}
\begin{tabular}{|l|l|l|l|l|l|l|l|l|l|l|l|l|}
\hline
& \multicolumn{6}{c|}{ Home A } & \multicolumn{6}{c|}{ Home B} \\ \hline
& \multicolumn{3}{l|}{Few-Shot \acronym{}}& \multicolumn{3}{l|}{Supervised baselines}& \multicolumn{3}{l|}{Few-Shot \acronym{}}& \multicolumn{3}{l|}{Supervised baselines}  \\ \hline
Activity            & K=3  &  K=5  & K=7  & {MLP} & \multicolumn{1}{l|}{\begin{tabular}[c]{@{}l@{}}BiLSTM\end{tabular}} & \multicolumn{1}{l|}{\begin{tabular}[c]{@{}l@{}}DeepConv\\ LSTM\end{tabular}} & K=3  & K=5  & K=7  & {MLP} & \multicolumn{1}{l|}{\begin{tabular}[c]{@{}l@{}}BiLSTM\end{tabular}} & \multicolumn{1}{l|}{\begin{tabular}[c]{@{}l@{}}DeepConv\\ LSTM\end{tabular}} \\ \hline
Leaving home        & 1.00 &  1.00 & 1.00 & {0.73}   & 0.71          & 0.97         & 0.87 & 0.87 & 0.88 & {0.62}   & 0.49          & 0.88 \\ \hline
Personal care       & 0.99 &  0.99 & 0.99 & {0.96}   & 0.92          & 0.98         & 0.96 & 0.96 & 0.96 & {0.83}   & 0.88          & 0.94 \\ \hline
Preparing breakfast & 0.70 &  0.88 & 0.87 & {0.87}   & 0.80          & 0.91         & 0.73 & 0.79 & 0.78 & {0.27}   & 0.19          & 0.33 \\ \hline
Preparing dinner    & -    &  -    & -    & -           & -             & -            & 0.29 & 0.33 & 0.36 & {0.02}   & 0.04          & 0.09 \\ \hline
Preparing lunch     & 0.92 &  0.96 & 0.96 & {0.95}   & 0.93          & 0.96         & 0.32 & 0.41 & 0.43 & {0.10}   & 0.18          & 0.08 \\ \hline
Relaxing on couch   & 0.85 &  0.91 & 0.96 & {0.99}   & 1.00          & 0.99         & 0.61 & 0.67 & 0.71 & {0.85}   & 0.84          & 0.85 \\ \hline
Showering           & 0.99 &  1.00 & 1.00 & {0.98}   & 0.93          & 0.98         & 0.95 & 0.95 & 0.96 & {0.95}   & 0.87          & 0.95 \\ \hline
Sleeping            & 0.92 &  0.94 & 0.97 & {1.00}   & 0.97          & 1.00         & 0.86 & 0.87 & 0.92 & {0.99}   & 0.86          & 0.99 \\ \hline
Snacking            & 0.27 &  0.48 & 0.57 & {0.24}   & 0.00          & 0.31         & 0.34 & 0.36 & 0.38 & {0.35}   & 0.29          & 0.46 \\ \hline
Weighted F1-score   & 0.91 &  0.95 & 0.96 & {0.95}   & 0.91          & 0.97         & 0.74 & 0.76 & 0.78 & {0.64}   & 0.63          & 0.72\\ \hline
\end{tabular}
\end{table}


\begin{table}[]
\tiny
\caption{MARBLE: Few-Shot \acronym{} vs the supervised baselines ($10\%$ of the training set)}
\label{tab:fsp_10_percent_marble}

\begin{tabular}{|l|l|l|l|l|l|l|}
\hline
 & \multicolumn{3}{l|}{Few-Shot \acronym{}}& \multicolumn{3}{l|}{Supervised baselines} \\ \hline
Activity            & K=3  & K=5  & K=7  & {MLP} & \multicolumn{1}{l|}{\begin{tabular}[c]{@{}l@{}}BiLSTM\end{tabular}} & \multicolumn{1}{l|}{\begin{tabular}[c]{@{}l@{}}DeepConv\\ LSTM\end{tabular}} \\ \hline
Clearing table      & 0.47 & 0.28 & 0.33 & {0.38}   & 0.00        & 0.46         \\ \hline
Eating              & 0.87 & 0.84 & 0.85 & {0.88}   & 0.80        & 0.90         \\ \hline
Entering home       & 0.65 & 0.69 & 0.61 & {0.00}   & 0.00        & 0.00         \\ \hline
Leaving home        & 0.71 & 0.86 & 0.80 & {0.00}   & 0.00        & 0.00         \\ \hline
Phone call          & 0.99 & 0.99 & 0.99 & {0.75}   & 0.63        & 0.79         \\ \hline
Preparing cold meal & 0.64 & 0.63 & 0.65 & {0.59}   & 0.43        & 0.58         \\ \hline
Preparing hot meal  & 0.80 & 0.83 & 0.87 & {0.81}   & 0.66        & 0.86         \\ \hline
Setting up table    & 0.27 & 0.38 & 0.45 & {0.16}   & 0.00        & 0.08         \\ \hline
Taking medicines    & 0.26 & 0.40 & 0.38 & {0.21}   & 0.06        & 0.35         \\ \hline
Using pc            & 0.97 & 0.98 & 0.98 & {0.90}   & 0.81        & 0.91         \\ \hline
Watching tv         & 0.98 & 0.99 & 0.99 & {0.96}   & 0.73        & 0.96         \\ \hline
Weighted F1-score   & 0.82 & 0.82 & 0.84 & {0.73}   & 0.58        & 0.76         \\ \hline
\end{tabular}
\end{table}

\FloatBarrier

\section{Examples of alternative prompts}
\label{alt_prompts_examples}

\begin{figure}[h!]
    \centering
    \includegraphics[width=0.8\textwidth]{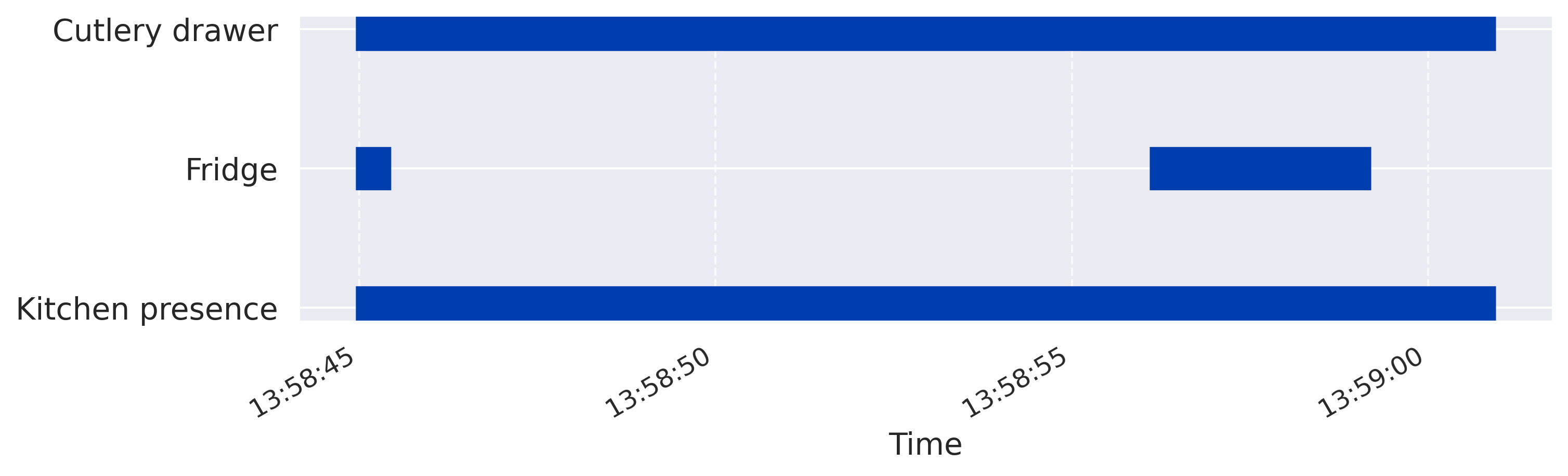}
    \caption{Visualization of the window used for the examples}
    \label{fig:window-example}
\end{figure}

In the following, we present a running example of a specific window and how it is processed by the alternative prompts considered in this paper. Figure \ref{fig:window-example} show the states of this window. Note that the presence in the kitchen and cutlery drawer are \textit{``already active''} and \textit{``persistent'' }states, the first segment of the fridge is an \textit{``already active''} state, while the second segment of the fridge is an \textit{``inner''} state.

\subsection{\acronym{}}
Here we show how the window in Figure~\ref{fig:window-example} is handled by the approach proposed in our paper.

\subsubsection{System prompt}\mbox{}\\ 
Act as a human activity recognition system that infers the activities performed by a subject in their home. The home has the following locations: 'dining room', 'hall', 'kitchen', 'living room', 'medicine area', 'office', and 'outside home'. The system captures the subject's interactions with chairs, couches, pantries, pots drawers, cutlery drawers, medicine cabinets, fridge, stove, and television. I will provide you with a description of what happened during a specific time window, and your goal is to provide the most likely activity. The possible activities are: 'phone call', 'clearing table', 'preparing hot meal', 'eating', 'entering home', 'leaving home', 'preparing cold meal', 'setting up table', 'taking medicines', 'using pc', and 'watching tv'. Your answer should be only one of these activities using the following format: \{'activity name'\}.

\subsubsection{User prompt}\mbox{}\\ 
During a 16-second time window (around 2:58 PM), the system observed the following. The subject is in the kitchen. Here, the fridge is already open, and the cutlery drawer is already open. After 1 second, they close the fridge. Then, they open the fridge. After 4 seconds, they close the fridge. What is the most likely activity performed by the subject? Your answer must be one and only one activity from the following activities: 'phone call', 'clearing table', 'preparing hot meal', 'eating', 'entering home', 'leaving home', 'preparing cold meal', 'setting up table', 'taking medicines', 'using pc', and 'watching tv'. Reason step by step.

\subsubsection{Output}\mbox{}\\ 
To determine the most likely activity performed by the subject during the observed time window, let's analyze the actions step by step:
\begin{enumerate}
    \item  \textbf{Location}: The subject is in the kitchen, which is relevant for activities related to food preparation or eating.
    \item  \textbf{Fridge Interaction}: The subject opens the fridge multiple times. The fridge is a common place to retrieve ingredients for meal preparation or to get food for eating.
    \item  \textbf{Cutlery Drawer Interaction}: The cutlery drawer is also open, indicating that the subject may be preparing to use utensils, which is typically associated with meal preparation or eating.
    \item  \textbf{Sequence of Actions}:
    \begin{itemize}
        \item The subject first has the fridge open and the cutlery drawer open, suggesting they are gathering items.
        \item They close the fridge after 1 second, which could indicate they retrieved something.
        \item They then open the fridge again, which suggests they might be looking for more items or confirming what they have.
        \item After 4 seconds, they close the fridge again, indicating they have finished gathering items.
    \end{itemize}
   \item  \textbf{Inference}: The combination of opening the fridge and cutlery drawer suggests that the subject is likely involved in preparing food. Given that the fridge is opened and closed multiple times, it indicates they are actively selecting items, which aligns with preparing a meal.
   \item  \textbf{Activity Options}: Among the possible activities, 'preparing hot meal' and 'preparing cold meal' are the most relevant. However, the actions observed (opening the fridge and cutlery drawer) do not specifically indicate the preparation of a hot meal, as there is no mention of cooking equipment like pots or the stove being used.
\end{enumerate}
Based on this analysis, the most likely activity performed by the subject is: \{'preparing cold meal'\}

\subsection{Cleland~\cite{cleland2024leveraging}}
Here we show how the \textbf{Cleland} strategy~\cite{cleland2024leveraging} handles the example in Figure~\ref{fig:window-example}. 
Consistently with the paper where this approach was presented, everything is in the user prompt, hence the system prompt is empty.

\subsubsection{System prompt}\mbox{}\\ 
-

\subsubsection{User prompt}\mbox{}\\ 
Sensors were placed around a home to monitor activity the following sentence describes which of these sensors have been triggered. It is 2:58:44 PM on 14-11-2018. Presence in the kitchen is detected. The fridge is closed. The fridge is opened. The fridge is closed. Out of the following activities, which one do you think was being performed taking time into account:'phone call', 'clearing table', 'preparing hot meal', 'eating', 'entering home', 'leaving home', 'preparing cold meal', 'setting up table', 'taking medicines', 'using pc', and 'watching tv'. The order of these categories does not matter. Respond using the following format: \{'activity name'\}.

\subsubsection{Output}\mbox{}\\ 
\{'preparing cold meal'\}

\subsection{ADL-LLM w/ absolute tss}
This approach differs from \acronym{} only in the user prompt, where it uses absolute timestamps instead of relative temporal relationships.

\subsubsection{System prompt}\mbox{}\\ 
Act as a human activity recognition system that infers the activities performed by a subject in their home. The home has the following locations: 'dining room', 'hall', 'kitchen', 'living room', 'medicine area', 'office', and 'outside home'. The system captures the subject's interactions with chairs, couches, pantries, pots drawers, cutlery drawers, medicine cabinets, fridge, stove, and television. I will provide you with a description of what happened during a specific time window, and your goal is to provide the most likely activity. The possible activities are: 'phone call', 'clearing table', 'preparing hot meal', 'eating', 'entering home', 'leaving home', 'preparing cold meal', 'setting up table', 'taking medicines', 'using pc', and 'watching tv'. Your answer should be only one of these activities using the following format: \{'activity name'\}.

\subsubsection{User prompt}\mbox{}\\ 
During a 16-second time window (starting at 2:58:44 PM), the system observed the following. The subject is in the kitchen. Here, the fridge is already open, and the cutlery drawer is already open. At 2:58:45 PM, they close the fridge. At 2:58:56 PM, they open the fridge. At 2:58:59 PM, they close the fridge. What is the most likely activity performed by the subject? Your answer must be one and only one activity from the following activities: 'phone call', 'clearing table', 'preparing hot meal', 'eating', 'entering home', 'leaving home', 'preparing cold meal', 'setting up table', 'taking medicines', 'using pc', and 'watching tv'. Reason step by step.

\subsubsection{Output}\mbox{}\\ 
To determine the most likely activity performed by the subject during the observed time window, let's analyze the details provided:

\begin{enumerate}
    \item  \textbf{Location}: The subject is in the kitchen, which is relevant for activities related to food preparation or eating.
    \item \textbf{Interactions}:
    \begin{itemize}
        \item The fridge is already open at the start of the observation.
        \item The cutlery drawer is also open, indicating that utensils are likely being accessed.
        \item The subject closes the fridge at 2:58:45 PM, which suggests they may have taken something out.
        \item The subject then opens the fridge again at 2:58:56 PM and closes it shortly after at 2:58:59 PM.
   \end{itemize}
    \item \textbf{Activity Analysis}:
    \begin{itemize}
        \item The repeated opening and closing of the fridge suggests that the subject is retrieving items from it, which is common in meal preparation.
        \item The open cutlery drawer indicates that the subject is likely preparing to use utensils, which further supports the idea of food preparation.
        \item The actions do not indicate eating, as there is no mention of sitting down or consuming food.
        \item There are no indications of other activities such as phone calls, clearing tables, or using a PC.
   \end{itemize}
\end{enumerate}
Based on this analysis, the most likely activity performed by the subject is related to preparing food, specifically a cold meal, as the fridge is involved and there is no indication of cooking or heating food.
Thus, the most likely activity is: \{'preparing cold meal'\}.

\subsection{ADL-LLM w/o CoT}
This approach differs from \acronym{} only in the user prompt, where the "Reason step by step" instruction is omitted.

\subsubsection{System prompt}\mbox{}\\ 
Act as a human activity recognition system that infers the activities performed by a subject in their home. The home has the following locations: 'dining room', 'hall', 'kitchen', 'living room', 'medicine area', 'office', and 'outside home'. The system captures the subject's interactions with chairs, couches, pantries, pots drawers, cutlery drawers, medicine cabinets, fridge, stove, and television. I will provide you with a description of what happened during a specific time window, and your goal is to provide the most likely activity. The possible activities are: 'phone call', 'clearing table', 'preparing hot meal', 'eating', 'entering home', 'leaving home', 'preparing cold meal', 'setting up table', 'taking medicines', 'using pc', and 'watching tv'. Your answer should be only one of these activities using the following format: \{'activity name'\}.

\subsubsection{User prompt}\mbox{}\\ 
During a 16-second time window (around 2:58 PM), the system observed the following. The subject is in the kitchen. Here, the fridge is already open, and the cutlery drawer is already open. After 1 second, they close the fridge. Then, they open the fridge. After 4 seconds, they close the fridge. What is the most likely activity performed by the subject? Your answer must be one and only one activity from the following activities: 'phone call', 'clearing table', 'preparing hot meal', 'eating', 'entering home', 'leaving home', 'preparing cold meal', 'setting up table', 'taking medicines', 'using pc', and 'watching tv'.

\subsubsection{Output}\mbox{}\\ 
\{'preparing cold meal'\}

\subsection{ADL-LLM w/o Env}
This approach differs from \acronym{} only in the system prompt, where the description of the monitored rooms and household items is omitted.

\subsubsection{System prompt}\mbox{}\\ 
Act as a human activity recognition system that infers the activities performed by a subject in their home. I will provide you with a description of what happened during a specific time window, and your goal is to provide the most likely activity. The possible activities are: 'phone call', 'clearing table', 'preparing hot meal', 'eating', 'entering home', 'leaving home', 'preparing cold meal', 'setting up table', 'taking medicines', 'using pc', and 'watching tv'. Your answer should be only one of these activities using the following format: \{'activity name'\}.

\subsubsection{User prompt}\mbox{}\\ 
During a 16-second time window (around 2:58 PM), the system observed the following. The subject is in the kitchen. Here, the fridge is already open, and the cutlery drawer is already open. After 1 second, they close the fridge. Then, they open the fridge. After 4 seconds, they close the fridge. What is the most likely activity performed by the subject? Your answer must be one and only one activity from the following activities: 'phone call', 'clearing table', 'preparing hot meal', 'eating', 'entering home', 'leaving home', 'preparing cold meal', 'setting up table', 'taking medicines', 'using pc', and 'watching tv'.

\subsubsection{Output}\mbox{}\\ 
\begin{enumerate}
    \item The subject is in the kitchen, which is a common location for meal preparation.
    \item The fridge is already open, indicating that the subject is likely retrieving ingredients or items for a meal.
    \item Opening and closing the fridge multiple times suggests that the subject is actively selecting items for their meal.
    \item The presence of the cutlery drawer being open could indicate that the subject is preparing a meal that requires utensils.
    \item Given the context of being in the kitchen, interacting with the fridge and cutlery drawer, and the repetitive opening and closing of the fridge, the most likely activity is \{'preparing cold meal'\}
\end{enumerate}

\bibliographystyle{ACM-Reference-Format}
\bibliography{references}